\documentclass[manuscript,nonacm]{acmart}

\AtBeginDocument{%
  \providecommand\BibTeX{{%
    \normalfont B\kern-0.5em{\scshape i\kern-0.25em b}\kern-0.8em\TeX}}}

\copyrightyear{2022} 
\acmYear{2022} 
\setcopyright{acmlicensed}\acmConference[FAccT '22]{2022 ACM Conference on Fairness, Accountability, and Transparency}{June 21--24, 2022}{Seoul, Republic of Korea}
\acmBooktitle{2022 ACM Conference on Fairness, Accountability, and Transparency (FAccT '22), June 21--24, 2022, Seoul, Republic of Korea}
\acmPrice{15.00}
\acmDOI{10.1145/3531146.3533206}
\acmISBN{978-1-4503-9352-2/22/06}

\begin{document}

\title[Should attention be all we need?]{Should attention be all we need? The epistemic and ethical implications of unification in machine learning}



\author{Nic Fishman}
\authornote{Both authors contributed equally to this research.}
\email{njwfish@gmail.com}
\affiliation{%
  \institution{University of Oxford}
  \city{Oxford}
  \country{UK}
}

\author{Leif Hancox-Li}
\email{leif.hancox-li@capitalone.com}
\authornotemark[1]
\affiliation{
  \institution{Capital One}
  \city{McLean}
  \state{VA}
  \country{USA}
  \postcode{22102}
}




\begin{abstract}
``Attention is all you need'' has become a fundamental precept in machine learning research. Originally designed for machine translation, transformers and the attention mechanisms that underpin them now find success across many problem domains. With the apparent domain-agnostic success of transformers, many researchers are excited that similar model architectures can be successfully deployed across diverse applications in vision, language and beyond. We consider the benefits and risks of these waves of unification on both epistemic and ethical fronts. On the epistemic side, we argue that many of the arguments in favor of unification in the natural sciences fail to transfer over to the machine learning case, or transfer over only under assumptions that might not hold. Unification also introduces epistemic risks related to portability, path dependency, methodological diversity, and increased black-boxing. On the ethical side, we discuss risks emerging from epistemic concerns, further marginalizing underrepresented perspectives, the centralization of power, and having fewer models across more domains of application.
\end{abstract}



\maketitle

\section{Introduction}
\label{unif}


In recent years, machine learning (ML) systems based on transformers have achieved new heights across multiple domains. Originally designed for translating languages \cite{vaswani2017attention}, they have since proved effective in other Natural Language Processing (NLP) tasks and in many other domains, from mainstays of ML research like computer vision (CV) \cite{dosovitskiy2020image} to newer arenas like autonomous driving \cite{prakash2021multi} and protein folding \cite{rao2020transformer}. Transformers do not just work, they dominate the leaderboards that measure progress\footnote{It is worth saying this ``progress'' is often fairly narrow; see \cite{raji2021ai} for a discussion of benchmarks/leaderboards and their limitations.} in ML: they are the top models for most NLP tasks \cite{sota2022translation, sota2022qa, sota2022sentiment}, but also in many tasks outside NLP. Transformers are the top 5 models for the largest image classification benchmark (Top 1 ImageNet accuracy) \cite{sota2022imagenet}, hold the top positions in image segmentation and object detection \cite{sota2022segment, sota2022object}, and top the charts in speech recognition \cite{sota2022speech}. Those successes cover the most active and well researched problems in ML, at least according to the number of benchmark submissions.

The empirical success of transformers beyond NLP has excited many researchers who have lauded the ``unification'' of different domains and tasks in ML. Where understanding NLP papers used to require specialized training in computational linguistics, CV researchers now feel that they can easily read papers across the sub-disciplinary divide, and vice versa, because of the shared underlying architecture of transformers \cite{twiml-nlp, twiml-vision}. Due to the newness of this phenomenon, statements about it tend to be made in less formal settings, industry blog posts/press releases \cite{deepai2021, scale2022, msft2022} and the popular press \cite{ornes2022}. But we feel it is fair to claim there is a sense in the community that 1) transformers offer a unified approach to ML problems and 2) unification is beneficial for ML.

The rise of transformers is related to ``foundation models'' as described in a Stanford workshop report \cite{bommasani2021opportunities}.
That paper defines a foundation model as ``any model that is trained on broad data at scale and can be adapted (e.g., fine-tuned) to a wide range of downstream tasks.'' Most ``foundation models'' identified in that paper are built using transformer architectures, but the two are conceptually distinct. A variational autoencoder could be a foundation model while not being a transformer; conversely AlphaFold, a transformer that predicts protein structure, is trained for a particular prediction problem rather than a general objective appropriate for fine-tuning for downstream tasks.

These differences aside, \citeauthor{bommasani2021opportunities} identify what we call ``unification'' (``homogenization'' in their terms) as a key feature of the history of ML as a field \cite{bommasani2021opportunities}. They pick out two discrete waves of unification. First, the rise of deep learning around 2015, pushing non-neural methods to the periphery of ML research (at least in CV and NLP) and eliminating the need to customize feature pipelines for each problem domain. This was carried out under the auspices of end-to-end learning which could learn good models from ``raw''\footnote{Although ``raw data'' is itself an oxymoron; see \cite{gitelman2013raw}.} feature inputs. The more recent development and promulgation of transformers across ML forms a second wave of unification, further accelerating these trends. Just as deep learning marginalized other general purpose ML methods (like boosting or kernel methods), transformers have further narrowed the scope of deployed ML methods to a specific neural architecture. The success of end-to-end learning went some way toward obviating the need for domain expertise in developing ML models. Transformers continue in this direction, unifying architectures across CV and NLP, fields that had formerly been dominated by convolutional neural networks (CNNs) and recurrent neural networks (RNNs) respectively. In both developments we can see two related phenomena: the unification of general purpose ML methods and the convergence of the expertise required to deploy these methods. 

In this paper, we take a critical eye to the excitement that ML researchers have expressed about unification being ``the'' way for ML to progress, highlighting the benefits and risks on both epistemic and ethical fronts. Our remit is less broad than what is covered in \cite{bommasani2021opportunities}. We focus on the unification aspect of this new methodology, leaving aside issues that are specifically due to models being large or trained on large amounts of data.\footnote{While most of the models that fall under our discussion are trained this way, we think that our points will apply even if they are trained on smaller datasets. See \cite{parrots} and \cite{bommasani2021opportunities} for analyses of the risks of training on big data in particular.}

While transformers are our motivating example for investigating unification, it is possible, or even likely, that transformers will not be the single method that unifies AI research; that may be something new that comes later. Even so, we believe that many of the arguments in this paper will apply to any unified approach that emerges out of a system like the current ML environment.\footnote{In fact, we believe many of the epistemic concerns we identify apply equally well to unification in any methodological field. We do not discuss these wider issues, but an interested reader can see \cite{cartwright2021rigour} for an account of the epistemic risks entailed in the social sciences' unification behind the causal inference paradigm.}

We start by considering the connection between unification and ``artificial general intelligence'' (Section \ref{agi}). These connections underpin some of the motivations underlying researchers' pursuit of unification. Then, we discuss the possible epistemic benefits (Section \ref{epistemic}), epistemic risks (Section \ref{erisks}), ethical benefits (Section \ref{etben}), and ethical risks (Section \ref{etrisks}) of unification.

\section{Unification and the pursuit of ``artificial general intelligence''}
\label{agi}

Unification is related to artificial general intelligence (AGI)\footnote{Also known as strong AI or human-level machine intelligence.} in two ways. The first is a more general pragmatic connection between unification and AGI, while the second is more ontological and particular to the current neural network paradigm in ML.

It has long been a staple idea of ML research that methods which are promising for AGI should be able to be successfully deployed in a variety of applications. Early work on AI was explicitly predicated on the idea that deploying a ``general purpose'' learning method on multiple tasks was evidence that the method could be extended to fully realize AGI \cite{rosenblatt1961principles, mcclelland1986parallel, papert1988one}. Although it is rarely as explicit in modern ML research, this lineage still animates a pragmatist vein of AGI optimism: if a particular general purpose learning framework is working well across all the narrow tasks the field can throw at it, then perhaps that framework is the one that will finally lead to the white whale \cite{mitchell2021ai, raji2021ai}.

A different perspective common in the field today is that, to quote a popular deep learning textbook, ``the brain provides a proof by example that intelligent behavior is possible, and a conceptually straightforward path to building intelligence is to reverse engineer the computational principles behind the brain and duplicate its functionality'' \cite{Goodfellow-et-al-2016}. As early as 2014 (before many of the most impressive deep learning advances) a survey found that almost half of ML experts believed that the route to AGI would be via cognitive science, and a further 39\% believed it would be specifically achieved by artificial neural networks \cite{muller2016future}. This view (implicitly) posits an ontological basis for intelligence, supporting the belief that the way to achieve artificial intelligence is to simulate that basis in relevant ways.\footnote{The metaphor of brain as computer may itself have ethical implications as well \cite{baria2021brain}, but we do not examine these here.}

This connection between unification and AGI is adjacent to the epistemic/ethical concerns we discuss below, which center around the benefits and risks of unification regardless of whether it leads to AGI.
We completely bracket the ethical benefits or risks of AGI, both due to the uncertainty about whether current approaches will bring about AGI \cite{lake2017building, shevlin2019limits} and in the interest of deferring to the large literature on the subject \cite{bostrom2017superintelligence, everitt2018agi, bostrom2018ethics}.
However, AGI is related to our discussion of the epistemic benefits of unification. On certain philosophical views, a shared ontology of learning systems (which underlies the connection between neural networks and AGI) presents real epistemic benefits. In Section \ref{ontology} we discuss whether the ontological motivation for pursuing AGI applies to today's transformer-driven trends of unification.

Unlike the ontological case, the pragmatic case for the connection between unification and AGI is essentially empiricist if we set aside questions about whether it will actually bring about AGI: it simply states that ML practitioners should pursue what works empirically. We address the epistemology of this in Section \ref{path} and \ref{diversity}. In doing so, we question how ``pragmatic'' this methodology actually is.


\section{Epistemic Benefits of Unification}
\label{epistemic}

Unification is often considered a virtue in science. Unification is a methodological driving force in physics, as demonstrated by the importance of the Standard Model in particle physics, or by the widespread influence of a few "fundamental laws" in diverse areas of physics.\footnote{We realize that there is disagreement over how unified physics actually is; see \cite{cartwright1999dappled} for a countervailing perspective. Our aim here is not to resolve this debate but to investigate the extent to which arguments for unification in science can apply to unification in AI.} 
Similarly in biology, Darwinism and later the Modern Synthesis were heralded partly because they provided unified accounts of diverse biological phenomena \cite{skipper1999selection, kitcher1981explanatory, plutynski2005explanatory}.
Philosophers of science have provided several reasons for why unification might indeed be a virtue \cite{sep-scientific-unity}.

In light of this, perhaps it's natural for AI practitioners to seek unified approaches to AI. But when we consider the trends towards unification in ML that were identified in Section \ref{unif}, 
it becomes less clear that unification in these senses is an epistemically desirable move. In what follows, we'll look at some general arguments for the desirability of unification in the sciences, and consider whether they transfer well to unification in ML.

A note before we proceed: we rely on many accounts of unification in science that are disputed by some philosophers. Resolving these disputes is beyond the scope of this paper. Our intent here is merely to pick out the few most prominent reasons for why we might want unification in science, and see if they apply to the types of unification in AI we consider. Of course, if these accounts of unification are fundamentally flawed, then that could shed doubt on whether they apply to unification in AI. Here we set aside the possibility of these fundamental flaws and merely consider if these accounts of unification can be transferred to the AI case, given the differences between AI and the natural sciences.

\subsection{Ontological unification}\label{ontology}

Unification is often seen as desirable in science because in many cases, we think that the disparate phenomena we want to unify share an ontological basis. Given this basis, it may be reasonable to expect a shared model describing how that basis generates these phenomena. One representative definition of ontological unification is as follows: ``Ontological unification is a matter of redescribing apparently independent and diverse phenomena as manifestations (outcomes, phases, forms, aspects) of one and the same small number of entities, powers, and processes'' \cite{maki2001explanatory}. For example, our knowledge that all matter is composed of the same finite set of elementary particles governed by four fundamental forces leads us to expect fruitful consequences from redescribing various phenomena in terms of these particles and forces.

Could ontological unification be a reason to pursue unification in ML? One reason it could be is if the deep learning architectures we discuss have a shared ontological basis with the human mind. In that case, we could redescribe both human minds and these ML systems as manifestations of the same underlying structures of real or simulated neurons. In other words, unifying ML systems could be a good epistemic move if we expect that the unified architectures represent the structure of the brain and if one of our goals is to create human-like artificial intelligence or AGI. To understand the appeal of ontological unification in ML, we'll next interrogate the extent to which current ``state-of-the-art'' artificial neural networks (ANNs) resemble the human brain.

The ANNs used in deep learning were initially developed in analogy to biological brains. ML researchers have claimed that they are similar to brains \cite{Goodfellow-et-al-2016, mcclelland2020placing}.  Many who write about the promise of deep learning do indeed lean on this idea that ever bigger and more sophisticated deep learning systems are the ``right'' systems to study in pursuit of AGI \cite{muller2016future}. Whether they are actually similar to brains and, more importantly, whether improvements in ANNs are driven by pushing ANNs to better approximate the functioning of brains, is a central point in the ontological case for unification.

In 1998 \citeauthor{o1998six} laid out six bare-minimum principles for a network to be considered biologically inspired \cite{o1998six}. These were 1) biological plausibility, 2) distributed representations, 3) inhibitory competition (there should be mechanisms limiting the number of neurons that can fire at once), 4) bidirectional activation propagation, 5) error-driven learning, 6) Hebbian learning (learning is the product of local, not global, information). At the time, ANNs satisfied only two of these (distributed representations and error-driven learning). Today ANNs have not moved much further in the direction of plausibility. They still lack inhibitory competition, they are feed-forward (not bidirectional) for the most part , and they (almost by definition) use global learning algorithms driven by globally defined loss functions. Perhaps more to the point, they are fundamentally not motivated by biological plausibility. Our intention here is to elaborate on this last point by examining some of the history of the microstructure \cite{mcclelland1986parallel} of ANNs to understand at various moments whether development turned on biological or pragmatic concerns. We will argue that the clear orientation toward ``what works'' and the differences between artificial and biological networks (both those outlined above and additional ones noted below) render the leap from ``human brains are AGI'' to ``ANNs can be AGI'' yet unwarranted. To be clear: it is possible that ANNs can be brought closer and closer in line with biology, and even possible that this will bring us closer and closer to AGI. We only intend to point out that such an outcome is not a certainty, nor is it even particularly strongly supported by current evidence.

The ANNs we know today are the descendants of the original perceptron algorithm \cite{rosenblatt1961principles}, which was based on a simplified mathematical model of a single neuron \cite{mcculloch1990logical, sejnowski2020unreasonable}. These earliest models used hard connectivity restrictions to maintain biological plausibility. Each neuron could only be connected to so many inputs, mimicking the physical constraints on biological neurons. The limitations of this model were at the heart of Minsky and Papert's original impossibility result, which proved that a perceptron could not compute various quantities without being fully connected to all inputs, that is, without being non-local (alternatively, the perceptron could be multi-layered, but this was dismissed as computationally intractable) \cite{minsky2017perceptrons, olazaran1996sociological}. Although the issues with multi-layered networks were eventually resolved, ANNs today are almost always fully connected in precisely the way that Minsky and Papert argued they would need to be, defying biological plausibility. 

In the 1980s two solutions emerged for training multi-layered perceptrons: first the Boltzmann machine learning algorithm \cite{ackley1985learning} and second the backpropagation approach \cite{rumelhart1986learning}. The Boltzmann approach had the benefit of being biologically plausible, the update step depending only on inputs and outputs of single neurons, which is in line with the Hebbian plasticity found in the neural cortex \cite{sejnowski2020unreasonable}. The backpropagation approach, on the other hand, is inconsistent with biological brains because it is a global update step. But backpropagation allowed for much more efficient training and was widely adopted, despite its deviance from biological plausibility. The ``non-locality'' of ANNs was mentioned above, but it bears repeating because it is probably the most central deviation from biological plausibility; central in the sense that it is so fundamentally structured into how practitioners work with deep networks. Backpropagation is the defining method that unites the vast array of methods now living under the ``neural'' umbrella: modern deep learning libraries are effectively auto-differentiation engines built to facilitate backpropagation \cite{tensorflow2015-whitepaper, NEURIPS2019_9015, jax2018github}.


Furthermore, while the original perceptron models certainly worked as analogues for biological neurons, they were greatly simplified. These simplifications are exemplified by two features of biological neural networks strikingly absent from their artificial counterparts: spiking behavior and dendritic non-linearities. In biological neural networks neuron activations are event-driven: the neuron is activated by spikes in its inputs, responding only when inputs change. On a practical level, spiking neurons are much more energy-efficient than contemporary ANNs, but this difference also means that the kind of computation that artificial and biological neural nets perform are fundamentally different \cite{roy2019towards}. Although spiking behavior has been known about and proposed as an alternative to standard neural nets since the 1990s, it has never really gained significant purchase in the ML community, in large part because it would be difficult to facilitate with standard computational hardware \cite{roy2019towards}.

Artificial and biological neurons also differ in their dendritic non-linearities. In deep neural nets a neuron takes a linear combination of its inputs and then applies the non-linearity, that is \tiny $f\left(\sum_i w^{(i)} x^{(i)}_{in}\right)$ \normalsize, but in biological neurons dendrites impose non-linearities on the inputs before summation, that is \tiny $f\left(\sum_i w^{(i)} g^{(i)}\left(x^{(i)}_{in}\right)\right)$ \normalsize \cite{jones2020can}. This change would be very straightforward to implement in modern ANNs, in many cases requiring changes to only a single line of code. But it has not seen widespread adoption because biological plausibility is not the main force animating deep learning research.

In contrast to perceptrons, the recent move to transformers had no biological inspiration at all.\footnote{Although CNNs, which previously dominated CV research, did stem from study of complex cells in the visual cortex \cite{fukushima1980self, richards2019deep}}. Rather, they developed out of a series of insights to improve language translation, the development of the attention mechanism to improve RNNs, and the subsequent realization that ``attention is all you need'' \cite{vaswani2017attention}. Transformers now regularly outperform the biologically-inspired CNNs on precisely the kind of CV tasks that CNN-like cells perform in the visual cortex \cite{chaudhari2021attentive}. This may be an artefact of path dependency in research (see Sec. \ref{path}), but if that were true it would only further validate our point in this section: ML researchers are not primarily interested in biological plausibility. If they are so not-motivated by biological plausibility that they abandon biologically inspired architectures capable of performing equally well without real cause, that may be symptomatic of problems in the field, but it certainly does not detract from the argument here. 

We do not intend this to be an exhaustive outline of differences between artificial and biological neural networks. Rather we have argued two things we think most ANN researchers would agree with. Firstly, we showed how ANN research is driven primarily by methodological pragmatism rather than biological plausibility. Secondly, we outlined how this pragmatism has pushed ANNs used in practice away from biology, rather than towards it, over time. Pragmatic reasons drove fully connected architectures, led to the adoption of backpropagation over Hebbian update rules, underlie the failure of spiking neurons to ever receive significant uptake, and are why transformers swept the research landscape despite a lack of any biological basis. We have not even touched on more fundamental\footnote{In the sense that architectural tweaks alone cannot fix them.} differences between deep learning and cognition, such as the non-causal, non-experimental and non-compositional nature of the most successful ML models \cite{lake2017building},\footnote{Reinforcement learning (RL) models can bridge this gap to some extent, but the RL models that achieve human-level performance invariably require architectures that are highly biologically implausible. For example, AlphaGo \cite{schrittwieser2020mastering} uses an explicit tree search to plan, which is decidedly non-biological.} the lack of ambiguity in AI \cite{birhane2021impossibility}, or that intelligence is not all in the brain \cite{mitchell2021ai}. 

To be clear, these limitations do not mean ANNs cannot be useful for investigating biological brains. Neuroscience has found some use in ANNs as a model \cite{richards2019deep, xiao2021training}. But using ANNs in that way requires careful construction and comparison to ensure meaningful inferences can be drawn precisely because of these (and other) disanalogies baked into the technology. Failure to account for this can lead to misleading conclusions and faulty science \cite{bowers2022deep}. On the other hand, nothing we have said means that ANNs cannot be brought further in line with biology to fruitful ends. Post-hoc biological analogies can point to promising avenues for taking the pragmatically successful aspects of modern ANNs and ``biologizing'' them,\footnote{See \cite{bricken2021attention}, which constructs such an analogy for transformers.} and if that process (or more generally, the process of bringing ANNs more in line with their biological counterparts) led to more performant models, it could form the foundation of an argument for a shared ontology. But that work has not yet been done, and so cannot (yet) be used to demonstrate a shared ontological basis for biological and artificial neural networks. As it stands, structural differences sufficiently distinguish  biological and artificial neural networks to make simplistic analogies (``ANNs are simulations of biological brains'') inadequate as grounds for a shared ontological basis.

In short, we do not think the current trajectory of unifying ML through transformers is well-motivated by a methodological principle of ontological unification, because the current top candidates for unification are too different from human brains and show no sign of coming closer to them.

\subsection{Explanatory unification}

Having addressed the ontological argument for unification, we now turn to more pragmatic reasons for unification.

One reason to pursue unification in science is that unification is itself explanatory, and explanation is one of the goals of science.
\citeauthor{friedman1974explanation} uses the example of the kinetic theory of gases as having ``reduced a multiplicity of unexplained, independent phenomena to one'' \cite{friedman1974explanation}. Where before we had separate empirical laws like the Boyle-Charles Law and Graham's Law, we can now use the kinetic theory of gases to understand why those laws and many others hold. This reduces ``the total number of independent phenomena that we have to accept as ultimate or given.''\footnote{Examples of explanatory unification may overlap with ontological unification, because a shared ontology is often thought to be explanatory. We separate ontological and explanatory unification here because we are leaving open the possibility that are types of explanatory unification that aren't identical to ontological unification.} Similarly, one might hope that instead of accepting as a brute fact that many model architectures work similarly well, having one or two model architectures that perform better than others would improve our understanding of AI.

We think that this reason to pursue unification does not apply to the types of unification we're addressing in this paper (at least right now). In particular, as we outline below, researchers struggle to explain how or why ANNs work in the settings where they are successfully deployed today. Thus explainability is not currently an epistemic benefit to the unification behind transformers (or ANNs before them), because although these models \textit{may} become explainable, that is by no means guaranteed. To make this clear, it is worth delineating the possible phenomena that this unification might explain.

\subsubsection{Possible explananda 1: commonalities between AI systems and human intelligence}

One group of candidate explananda is the alleged similarities between AI and human intelligence. The thought is that if the human brain has a similar structure to some ANN that's successful across a broad range of tasks and domains in AI, that could be explained by the hypothesis that the structure of that neural network is the key to AGI.

Whether these are valid explananda for the AI architectures in question depends on the issues we discuss in Section \ref{ontology}. As we argue there, we do not think that current neuroscientific evidence supports the hypothesis that the current crop of allegedly general-purpose neural network architectures resemble the brain's architecture.

It's possible that one day we will have a unifying AI architecture that better resembles what we know of the brain's structure. If models with that architecture prove to resemble human intelligence in their performance, then that architecture may explain why humans and those AI systems are similarly intelligent. But as things currently stand, the architectures being pursued in the name of unification will not be able to explain these hypothetical explananda.

\subsubsection{Possible explananda 2: inductive biases}

The structures of ANNs encode particular inductive biases that allow them to perform well across different tasks. This points to a second possible type of explanatory unification: a unified explanation for these inductive biases. In the ML literature, ``inductive biases'' refers to ``preferences'' for the kinds of functions a model represents \cite{goyal2020inductive}. 
For example, ridge regression has an inductive bias for the coefficients to be closer to zero, and the strength of this bias can be tuned by a regularization parameter $\lambda$. The No Free Lunch theorem in ML \cite{wolpert1997no} essentially states that all ML algorithms will encode some such set of inductive biases, and that these preferences over the function space are necessary for ML models to generalize to new data. Neural networks clearly encode some kind of inductive biases, and these biases are almost by definition central to their ability to generalize to new data. If the full array of inductive biases of ANNs could be outlined, it would go a long way toward explaining how and why AI models work. Unified models could help provide a common basis for these shared inductive biases. This area has been of central importance to the literature on the theory of deep learning for a number of years \cite{neyshabur2014search}, but deep learning is a fundamentally empiricist field and the theory tends to lag far behind practice. As it stands today, deep learning theory cannot outline the inductive biases of simple multi-layer perceptron neural networks trained with stochastic gradient descent, much less give an account of the sophisticated architectures achieving state of the art results. Ongoing work on this project is promising as a way forward for the field \cite{goyal2020inductive}. If an account of the inductive biases induced by different architectural choices could be constructed, that would go a long way toward arguing for a unified ANN architecture on the basis of explanatory unification. 

\subsubsection{Possible explananda 3: successful performance by one system across different tasks and modalities}
\label{many-models}

Another possible class of explananda that unified models may explain is the mechanisms underlying successful prediction across different tasks and modalities. We suspect that this is part of why the expanded success of transformers across modalities is so exciting to many researchers: they hope that multi-modal, multi-task models are sharing information across different modalities and tasks. Similarly, there is a hope of shared mechanisms: if what we think transformers are doing with sequential information in language is also what they're doing with image patches engineered to be sequential, then we can understand predictive success in multiple domains using a similar mechanism.

\begin{figure}
    \centering
    \includegraphics[width=\columnwidth]{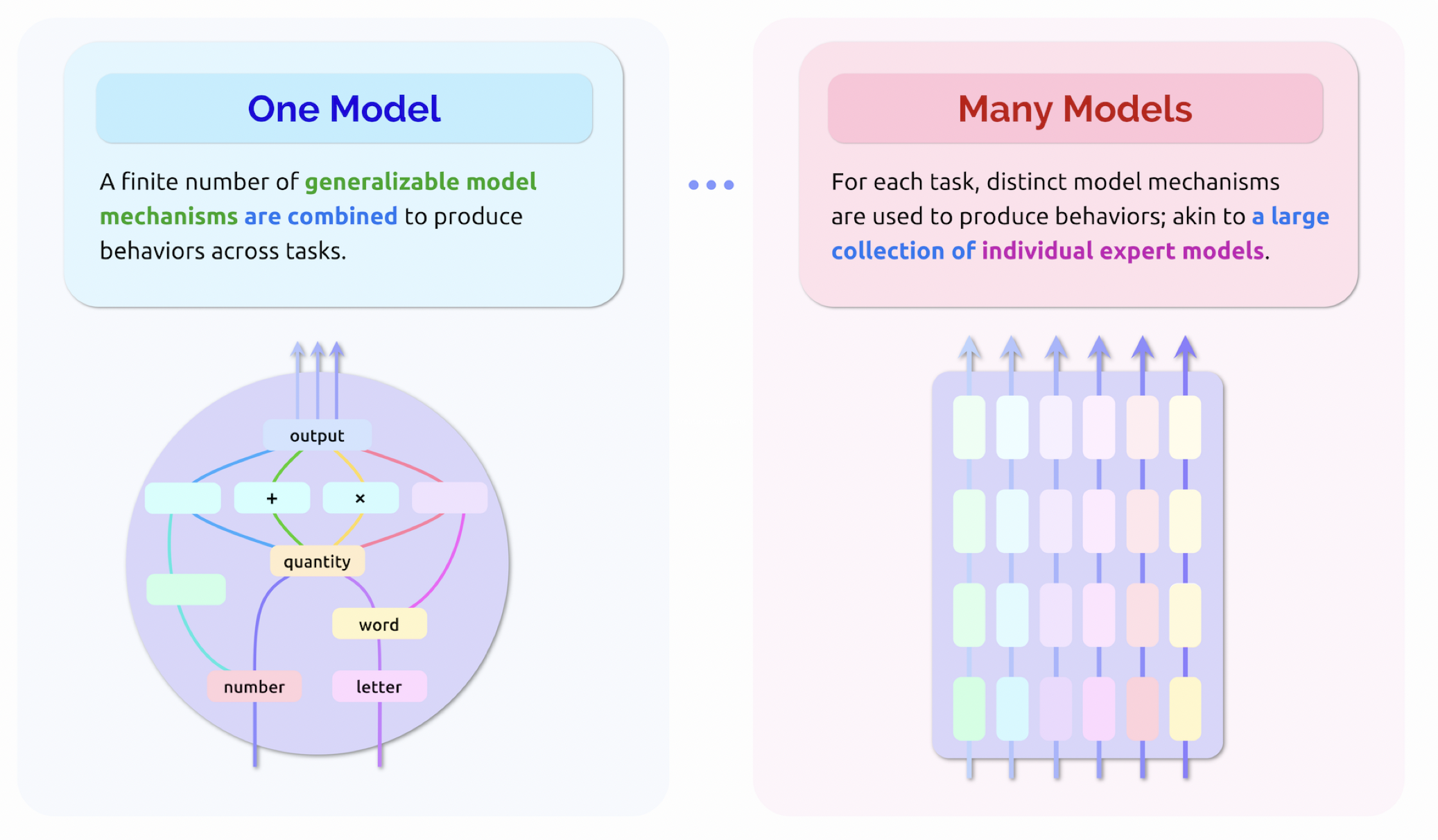}
    \Description{Diagram depicting one model with a few mechanisms on the left, and many combined models with many mechanisms on the right.}
    \caption{Unified models may fall on a spectrum from having a few mechanisms that generalize to many modalities and tasks (left hand side), or to the other extreme of having separate mechanisms for each task they can perform (right hand side).  Figure by \citeauthor{bommasani2021opportunities} \cite{bommasani2021opportunities}, licensed under \href{https://creativecommons.org/licenses/by/4.0/}{CC BY 4.0}.}
    \label{fig:ommm}
\end{figure}

It is as yet unclear if current trends towards unification will bring about this type of unification-through-shared-mechanisms. As \citeauthor{bommasani2021opportunities} point out, unified models fall on a spectrum between one model with shared mechanisms for many tasks, or many models that just happen to be ``glued'' together into one large model (Figure \ref{fig:ommm}):
\begin{quote}
    As one model, behavior can be made interpretable by identifying and characterising the finite number of generalizable model mechanisms that the model uses to produce behaviors across tasks (e.g., mechanisms that assign meaning to words, compare quantities, and perform arithmetic). As many models, explanations of model behavior in one task are not necessarily informative about behavior in other tasks; thus requiring [sic] to study behavior independently in each task \cite{bommasani2021opportunities}.
\end{quote}

Given this, if unified models are more like the ``one model'' end of the spectrum, then they would be better at explaining the explananda described here. If they are more like the ``many models'' end of the spectrum, then these explananda would not find a unified explanation within the model. Given the newness of these unified models and the difficulties we have in understanding how they make decisions \cite{vashishth2019attention}, it is unclear at present where on the spectrum they lie.

\subsubsection{Review of explanatory unification arguments}

In short, the prospects for explanatory unification through the type of architectural unification we discuss in this paper are mixed. For now, it is unclear if unification would be explanatory of intelligence as a phenomenon, of shared inductive biases, or of success across multiple tasks and domains. As the field develops and we get better insight into the inner mechanisms of the unified models, we will hopefully get a clearer picture of how far explanatory unification can be a reason to prefer a unified model landscape. 

\subsection{Unification as parsimony}

In addition to unification being itself explanatory, philosophers of science have also put forward unification as a methodological principle that helps us achieve the epistemic value of simplicity or parsimony. The thought is that ``simpler hypotheses, models or theories present a higher likelihood of truth, empirical support and accurate prediction \cite{sep-scientific-unity}.'' This thought can be cashed out in different statistical frameworks \cite{sober2003two, myrvold2003bayesian},
but we will make some general points that do not assume any particular statistical framework.
In this view of unification-as-parsimony, having unified models explaining diverse domains of phenomena is simpler than having many different models for each domain.

How does ML unification fare under this argument? The verdict is mixed here. On the one hand, unification does lead to fewer models covering the same phenomena, and this could be viewed as a simplification. On the other hand, this unification is typically achieved by using more complex models than the counterfactual domain-specific models. The more domains a model has to cover, the more complex it typically is. The complexity consists not just in its internal mechanisms, but in the shape of the predictive function.
It's unclear how the unification-as-parsimony arguments would assess this trade-off, as the canonical examples of unification they discuss in science don't have this trade-off. This issue is also related to the one-model-many-models spectrum discussed in Section \ref{many-models}. To the extent that unified models are near the end of the many-models spectrum, they are likely to be less parsimonious.

A more fundamental problem with the unification-as-parsimony argument is that it's unclear how ML models of the type we're discussing relate to goals like ``truth'' and ``empirical support''. These models are selected on the basis of accurate prediction on a held-out test set, so by definition they already satisfy the goal of accurate prediction on past data. The emphasis on their utility as a tool for prediction makes it hard to understand what their relationship to ``truth'', for example, is. In contrast, models in the natural sciences are often aimed at revealing some true underlying mechanism of nature.
It is relatively intelligible to say of Darwinism, for example, that it is true, or that it depicts the truth about some phenomena, but it's unclear what it would mean to say that a transformer-based ML model depicts the truth.\footnote{We are not endorsing the controversial thesis of scientific realism in this claim--we ourselves take no position on whether scientific theories can be said to be true. We make this point to address audiences who see truth as one of the aims of scientific model-building, since one of the arguments for unification-as-parsimony appeals to the allegedly greater likelihood of simpler models being true.} Given differences like this between ML and the natural sciences, the unification-as-parsimony argument may need to be modified before it can be intelligibly applied to ML. 

\subsection{Unity of tools}

While we have been somewhat skeptical of various arguments for unification presented above, we do not want to deny the real epistemic advantages that unification can offer, which ML researchers have gestured to in their commentary on the phenomenon. Researchers are excited that they can now more easily read ML papers in other domains and transfer techniques useful in one domain to another domain \cite{twiml-nlp, twiml-vision, karpathy}. These are not trivial benefits. When we consider transformers as shared tools that are easily imported between contexts, they can be thought of as ``trading zones'' or ``boundary objects'' as defined by historians and sociologists of science \cite{galison1997image,star1989structure}. Both material and abstract objects can provide this kind of tool-based unity. In the ``trading zones'' and ``boundary objects'' framings of shared tools, a key idea is that shared tools provide a common ground for communication between collaborators, drawing on each collaborator's local understanding without necessitating global agreement. Insofar as we see the facilitation of such collaborations as an epistemic good, we can view transformers as providing epistemic benefits.

\section{Epistemic Risks}\label{erisks}

Having previously considered the possible benefits of having unified approaches to model-building across multiple tasks and domains,
we now consider some possible epistemic risks introduced by greater model unification.

\subsection{Turbocharging the tendency to ignore domain experts}\label{domain}

One epistemic disadvantage from having fewer, more unified models is the increased risk of falling into what \citeauthor{portability} call the ``Portability Trap'': ``Failure to understand how re-purposing algorithmic solutions designed for one social context may be misleading, inaccurate, or otherwise do harm when applied to a different context'' \cite{portability}.

This trap has led to some high-profile ML failures \cite{kelly2019key, moderation, noble2018algorithms}. 
As \citeauthor{portability} point out, the Portability Trap arises from the modularity taught as good computer programming practice: creating programs in a modular way so that modules can be easily applied to instances outside the originally intended domain.

Another way to characterize the trap is that it's generally a good idea to design a model with the help of domain experts, who can tell you what data to use, how to encode the data as features, how to detect when outputs are not what they should be, and so on. With the advent of unified models, the temptation to ignore domain experts, for cost reasons or otherwise, increases. After all, the model is supposed to work in multiple domains, modalities, and tasks, seemingly obviating the need to customize it to particular contexts.

Ignoring domain experts can also lead to a particular type of ethical risk in which the viewpoints and knowledge marginalized groups are unjustifiably ignored. We discuss this particular ethical risk in Section \ref{margin}.

\subsection{Path dependency}
\label{path}

Enormous amounts of time have been invested into software infrastructure like Tensorflow and Pytorch. Researchers have carefully refined skills and intuitions about ANNs and how to get them to work well. The volume of published findings about ANNs and the excitement around the area dwarfs that of any other subject in ML. 
All of this means that investigating other existing paradigms is structurally disincentivized: the lack of infrastructure means developing new models is more costly, the lack of intuition leads to a lack of ideas and frustrations in implementation, and the lack of excitement and new results means investments are unlikely to be rewarded. Put another way, the ``switching cost'' is extremely high. 

Additionally, the particular path ML has gone down is in many senses arbitrary: it is the product of hardware lotteries and the ever increasing availability of compute \cite{hooker2021hardware} combined with norms that push for state of the art results on metrics that are riddled with problems \cite{raji2021ai}. It would be one thing if these concerns were speculative, but there is evidence as well as a sense within the ML community that these are real problems.  Within ANN research, there are clear cases of path dependency where technologies that work are abandoned or pursued for arbitrary reasons.  One such case discussed by historians is that of over-specialization in chess-playing \cite{ensmenger2012chess}. Another example is where the combination of incentives to publish novel methods, expanding computational power, and a lack of incentives to fine-tune baselines leads to older, simpler methods being outcompeted by newer methods trained and fine tuned with larger compute budgets. This is clearly what happened in metric learning, where over a decade very little progress was made after accounting for increased compute \cite{musgrave2020metric}. In a different but related vein, transformers have found huge success in CV, but  some of this is due to engineering innovations in deep architecture design that can be implemented in many kinds of ANNs; importing  some of these architectural innovations into convolutional ResNets leads them to be competitive with transformers on large image datasets \cite{liu2022convnet}. 

These two facts---the high switching cost and the arbitrariness of the current trajectory of research---mean that the ML community, like almost all social communities \cite{liebowitz1995path, mahoney2000path}, is extremely path dependent.


In social contexts, path dependency often leads to sub-optimal social outcomes \cite{mahoney2000path}, which are of immediate ethical concern. In ML, path dependence is an epistemic concern if the community fails to fully explore alternative ways of developing machine learners. It may be useful to think about path dependence in the context of the explore/exploit trade off \cite{auer2002finite, devezer}, 
which philosophers of science have found useful in thinking through epistemic communities \cite{thoma2015epistemic, grim2019modeling}. Path dependence compounds the dilemma posed by the explore/exploit trade off by lowering the cost of exploiting each time the decision is made to exploit. Developing a Bayesian method to compete with ANNs on image classification (for example) is risky in the classic explore/exploit sense in that it might not work, but the dice are loaded: the comparative lack of infrastructure, intuition, and community make it extremely difficult to develop a model of comparable sophistication in a non-deep learning paradigm. In the end, the only way to know whether alternative approaches to ML would be more effective would be to invest similar amounts of time and energy into those alternatives. We take this to be a central epistemic risk of any methodologically homogeneous community.

\subsection{Methodological Diversity in the sciences}
\label{diversity}

ML research is at least partly preoccupied with producing methods that can be applied to generate knowledge in other fields, from physics and biology to economics and history. The importance of these ``epistemic spillovers'' to the epistemic value of ML research creates another key risk in the unification of ML. 
In scientific enterprises, deriving the same answer from multiple different methodologies is an important tool for confirming hypotheses. This is known as triangulation, and it leads to more robust scientific conclusions
\cite{heesen2019vindicating}. Importantly, this result holds even if one is unsure which methods can be trusted. Such a powerful approach can undoubtedly be very helpful in the difficult non-parametric settings where ML is often required. When ML becomes more unified, that crowds out alternative methods (see Sec. \ref{path}) which undermines the possibility of triangulation. If there are not enough methods that work sufficiently well to ask the relevant questions, then the answers cannot be triangulated to give the added confidence in the conclusions.

Another argument for methodological plurality comes from research on cognitive diversity \cite{hong2004groups}. This line of research argues that a group of diverse problem solvers are (under a particular set of assumptions) more effective than a group of individually effective problem solvers. These arguments can be extended to epistemic communities \cite{zollman2010epistemic}, although some caution is required. In particular, \citeauthor{grim2019diversity} note that the results proved in \cite{hong2004groups} show that under certain circumstances, a group of diverse problem solvers (models) \emph{can} outperform a homogeneous group of individually effective problem solvers \cite{grim2019diversity}. They also demonstrate that true ``experts'' (as opposed to just more-effective problem solvers) can be substantially better than a diverse group of non-experts. We agree that expert input in the form of domain knowledge is often crucial, especially in science. But that argument cuts against the premise of general purpose ML tools, which tend to push domain experts out of the process (see Sec. \ref{domain}). 

The diversity results in \cite{hong2004groups} apply equally to ML methods: a diverse ensemble of models can more effectively solve problems than an ensemble of individually highly effective models. This fact has also been noted independently within ML where diversity of base classifiers empirically improves boosting \cite{melville2004diverse} and diverse deep ensembles empirically improve performance and uncertainty quantification \cite{lee2015m}. The diversity version of the argument for methodological pluralism more directly responds to the pragmatic argument for ML unification (``do what performs well on benchmarks''): first, if we consider ensembles, diverse base models empirically perform well relative to homogenous base models; second, with respect to the field's epistemic spillovers, a diverse array of minimally viable methods is likely epistemically preferable to a single best method. 

Combining the benefits of diversity with the confirmatory basis afforded by triangulation, there is a strong epistemic case against unification in ML. Both of these arguments rely on the existence of alternatives, but these alternatives must meet a bare threshold of viability. By viability we mean that alternatives must be flexible enough to accommodate the kinds of research questions practitioners will seek to answer, computationally efficient enough to return answers in a reasonable time frame, and refined enough for those answers to be meaningful (at minimum providing better-than-random predictions). Considering the problems raised by path dependency in ML, it is clear how the issues discussed in Section \ref{path} compound these concerns about depriving non-ML researchers of methodological pluralism.

\subsection{Increased black-boxing}
\label{black-box}

The drive to unification will, at least at first, have a tendency to increase the opaque nature of the resulting models. Referring back to the one model-many models concept introduced in Section \ref{many-models}, unification often occurs by making an existing black-box ANN more complicated. The following moves made in the service of unification perpetuate this black-boxing tendency:  
\begin{enumerate}
    \item Eliminating feature engineering makes the model less understandable to humans, because feature engineering is often a stage where human-understandable features are created out of data that is less human-understandable.
    \item Many of the more unified multi-modal, multi-domain models come about by combining model architecture ideas inspired by different modalities. How the resulting model generates outputs becomes even more opaque than in the original unimodal models that inspired it, because information from different modalities is combined in ways that we don't fully understand.
\end{enumerate}

Against this, it must be noted that advances in model explainability might ameliorate some of these concerns. But it is by no means guaranteed that such advances will occur or would improve explainability enough to compensate for the model's increased complexity. Explainability for ANNs remains a contested issue. Criticisms of well-known explanation methods abound, and the criteria for what counts as a successful explanation are still being debated \cite{ghassemi2021false, ghorbani2019interpretation, gilpin2018explaining}.

Increased black-boxing also has ethical consequences distinct from the potentially higher risk of model failure. In many contexts, model explainability may be considered to be an ethical obligation, particularly when making high-stakes decisions. This is the intuition behind explainability requirements in laws like the GDPR \cite{felzmann2019transparency}.
Increased black-boxing in these contexts, then, would be not just an epistemic risk, but an ethical risk, as finding accurate explanations for ANNs is much more difficult than for, say, generalized additive models.

\section{Ethical Risks and Benefits of Unification}
\label{ethics}

\subsection{Ethical Benefits}\label{etben}

While subsequent sections will cover the many ethical risks that further architectural unification in ML presents, it's worth considering some possible ethical benefits that might result, given the appropriate socio-economic conditions. 

One benefit that has arguably resulted from the prevalence of widely used open source software (OSS) in all kinds of software applications is that many people gain access to the features of a shared software library. This claim is contestable, since one might argue that society would be better off without any of the software that rely on widely-used open source packages, or that one could have developed similarly good software without those open source packages. 

We do want to indicate the possibility, though, that ANNs with unified architectures could be managed as OSS, open to anyone to use and contribute to. Hugging Face is an example of a company that claims to ``democratize \emph{good} machine learning'' (emphasis theirs) by making ANN models open source \cite{hugging}. In some conceptions of how society can and should work, this could lead to net social benefits. However, the details around how to manage such unified AI systems to avoid unintended negative consequences are still under-specified. For that reason, we list this as a \emph{possible} benefit from unification, not an actual one.

Similarly, in analogy with OSS, if AI models were open sourced they could be assessed for vulnerabilities or audited for discriminatory biases. This is the ``many eyeballs'' model of security, and it has certain additional advantages when considering ML systems \cite{gebru2018datasheets, arnold2019factsheets}. The benefit comes with an attendant risk: if a vulnerability does escape the notice of many eyes, then the fact that one centralized package is used by many different people can become a security issue of staggering scale. The log4j vulnerability \cite{log4j} is a recent example, but within cybersecurity this tradeoff has been debated for decades \cite{hissam2002trust}. 

In short, we think having fewer, more unified models could have some ethical benefits with the appropriate social and institutional structures, but it's unclear if these benefits would accrue in the actual world.

\subsection{Ethical Risks}\label{etrisks}

The risks of large models have  recently been discussed extensively \cite{bommasani2021opportunities, parrots}.
Some of the points we make here will overlap with those discussions, because the model types we're considering overlap with the model types discussed in those papers. However, we focus particularly on the risks emanating from the unification of models.

\subsubsection{Ethical Implications of Epistemic Risks} \label{epieth}

ML systems are deployed in the real world and make decisions that affect the lives of billions of people every day, from whose social media posts get algorithmically boosted to which results appear on the first page of search engine to the automatic annotation of personal photos. This means that many of the epistemic risks we outline in Section \ref{erisks} have immediate ethical implications. On a basic level, the tendency to ignore domain experts (Section \ref{domain}) and the issues around path dependency (Section \ref{path}) may make ML systems less effective tools, which for systems that are so widely used has immediate social welfare implications.\footnote{Although if those systems are designed to operate against the best interests of their decision subjects, their suboptimality may be an ethical silver lining.}
And as we discussed in Section \ref{black-box}, the trend of increased black-boxing itself has distinct ethical implications related to subjects' rights to explanations.

\subsubsection{Making it easier to ignore marginalized perspectives}
\label{margin}

As we argued previously in Section \ref{domain}, model unification can amplify the existing tendency to ignore domain experts. This in turn adds to an ongoing pattern of building models without getting input from marginalized perspectives that could have informed key decisions about the model. We think this is not just an epistemic risk, but also an ethical risk. ML has a track record of disproportionately harming marginalized groups, 
and many critics have pointed out that the lack of diversity among the builders of ML probably doesn't help \cite{mclennan2020ai, diverse, ainow2017}.

When a system could harm marginalized groups, it is reasonable to think of members of these groups as domain experts on the decisions made by the system. When domain experts are left out of processes like determining what features are valid ones for predictions, these marginalized perspectives will also be left out. End-to-end learning is designed to cut humans out of the process of determining what features are good ones for the model to rely on. Having a plug-and-play model that can be easily used in different domains will put more power in the hands of ML engineers and adjacent roles, while taking power away from domain experts.

This will also abstract the systems further from their social context. If any engineer can take a CV model and plug-and-play it into an NLP setting, there is less friction in this process to encourage the engineer to consult a language expert. One ethical repercussion is that perspectives from sociolinguistics about language-related harms will be more likely to be left out.

This is related to the feminist epistemological values of considering context and multiple points of view \cite{haraway1988situated, harding1995strong}.
If we think of each model as giving us a certain perspective into its domains, feminist epistemologists favor having multiple such perspectives, since each perspective is influenced by its own particular background. There is no one ``best'' perspective. Models, like perspectives, embed particular social values, and the fewer models we have, the more a few privileged perspectives will get to determine the shape of the world. Having one or even just a few general purpose model architectures that supposedly can ``understand'' diverse domains is eerily close to the notion of having a ``view from nowhere'' \cite{nagel1989view}, which feminist epistemologists have criticized as privileging the white masculine gaze over more marginalized perspectives.

\subsubsection{Further centralization of power}

The increasing centralization of power in AI has been discussed by various activists and researchers, especially as it pertains to large ML models \cite{whittaker2021steep}.
There is an additional aspect of unification that lends itself to centralized power beyond the fact that these models are large. Unification expands the potential applications of a single model over many more areas of life. The vision that some researchers have is to be able to train a model on, say, a language corpus and then have it perform tasks in CV, knowledge-based reasoning, and so on. This ambition, to be general-purpose across different modalities, tasks, domains, and features, means that whoever controls a pre-trained model of this general purpose nature would wield broader power over more areas of human life.

\subsubsection{The ``algorithmic leviathan'' argument against algorithmic monocultures}

Another ethical risk of having a more unified model landscape is that \emph{systematic} arbitrariness in decision making can be morally wrong in a way distinct from any wrongs that might be attributed to arbitrariness in individual decisions.  \citeauthor{creel2021algorithmic} argue that arbitrariness in individual decisions may be morally problematic sometimes, but in situations where the arbitrariness does not violate fundamental rights, there could be an instrumental justification for arbitrariness \cite{creel2021algorithmic}. In contrast, arbitrariness at scale, where a single model is able to make arbitrary decisions that affect many people in many domains, introduces an element of wrongness over and above any wrongs wrought by individual arbitrary decisions. With arbitrariness at scale, an individual may face an arbitrary decision from an algorithm not just in one instance, but in many instances across many opportunities. This systematic exclusion is morally problematic under many moral theories: it could translate to a lack of capacities, lack of genuine freedom, or being at the bottom of a social hierarchy of esteem or domination \cite{sep-capability-approach, raz1986morality, anderson1999point}. Model unification increases the systematicity of arbitrariness, because it enables fewer models to make decisions over more domains.

\subsubsection{Epistemic homogeneity's impact on social welfare}

In Section \ref{path}, we put forward some epistemic considerations against narrowing the path of ML research to be more focused on unified models. In addition to these potential epistemic drawbacks, emerging research suggests that there could also be negative impacts on aggregate social welfare if the models used by organizations to make decisions are more accurate but more homogeneous.   \citeauthor{Kleinberge2018340118} argue that introducing a more accurate algorithm that's used by many organizations can drive society into an equilibrium that's worse than having many less accurate algorithms \cite{Kleinberge2018340118}. This leads to the average quality of decisions across society being lower, even if it makes sense for individual organizations to  each opt for the most accurate algorithm. \citeauthor{Kleinberge2018340118} make their point using the case of ranking algorithms, so the implications for other types of algorithms are unclear, but their work shows that we should be wary of assuming that having everyone use the same ``state of the art'' algorithm is necessarily better for society.

\section{Conclusion}

In this paper, we characterized key trends in recent AI research that unify model architectures across different modalities and tasks. We considered the potential benefits and risks of these trends on both epistemic and ethical fronts. We've focused specifically on the features of these models that are directly related to unification---that is, directly related to the aspiration to provide model architectures that can be applied across multiple domains, tasks, and modalities with little to no adjustment. This provides a complementary perspective to previous work that has discussed the implications of other properties of these models, such as their large size and their reliance on large corpuses of training data \cite{parrots, bommasani2021opportunities}.

We argue that on current evidence, the epistemic benefits are not as strong as analogous benefits from unification in the natural sciences. We discuss multiple epistemic risks arising from having too homogeneous a methodological community. Finally, we discuss possible ethical benefits from open-source unified models, and ethical risks like further marginalizing underrepresented perspectives and facilitating centralized, homogeneous decisions.

Moving forward, we think there are further possible consequences of unification that are worth exploring. For example, the performativity of ML models has been noted by other commentators---models have the potential to change the behavior of decision subjects, with this changed behavior in turn potentially reinforcing the models' ``correctness'', and so on in a positive feedback loop \cite{kockelman2020epistemic, hu2018short, bloor1996idealism}.
Do more unified models strengthen this dynamic where it exists, or make such feedback loops more probable? What about the phenomenon of ``Holy Grail performativity''---where the presence of an overarching goal of doing well across multiple predefined tasks, modalities and domains changes researcher behavior to be oriented around this quest, in a way that potentially forecloses other possibilities \cite{varshney2019pretrained}? We hope that these and other possible consequences of a more unified model landscape will be explored in future work.


\bibliographystyle{ACM-Reference-Format}
\bibliography{sample-base}


\begin{thebibliography}{116}


\ifx \showCODEN    \undefined \def \showCODEN     #1{\unskip}     \fi
\ifx \showDOI      \undefined \def \showDOI       #1{#1}\fi
\ifx \showISBNx    \undefined \def \showISBNx     #1{\unskip}     \fi
\ifx \showISBNxiii \undefined \def \showISBNxiii  #1{\unskip}     \fi
\ifx \showISSN     \undefined \def \showISSN      #1{\unskip}     \fi
\ifx \showLCCN     \undefined \def \showLCCN      #1{\unskip}     \fi
\ifx \shownote     \undefined \def \shownote      #1{#1}          \fi
\ifx \showarticletitle \undefined \def \showarticletitle #1{#1}   \fi
\ifx \showURL      \undefined \def \showURL       {\relax}        \fi
\providecommand\bibfield[2]{#2}
\providecommand\bibinfo[2]{#2}
\providecommand\natexlab[1]{#1}
\providecommand\showeprint[2][]{arXiv:#2}

\bibitem[Abadi et~al\mbox{.}(2015)]%
        {tensorflow2015-whitepaper}
\bibfield{author}{\bibinfo{person}{Mart\'{i}n Abadi}, \bibinfo{person}{Ashish
  Agarwal}, \bibinfo{person}{Paul Barham}, \bibinfo{person}{Eugene Brevdo},
  \bibinfo{person}{Zhifeng Chen}, \bibinfo{person}{Craig Citro},
  \bibinfo{person}{Greg~S. Corrado}, \bibinfo{person}{Andy Davis},
  \bibinfo{person}{Jeffrey Dean}, \bibinfo{person}{Matthieu Devin},
  \bibinfo{person}{Sanjay Ghemawat}, \bibinfo{person}{Ian Goodfellow},
  \bibinfo{person}{Andrew Harp}, \bibinfo{person}{Geoffrey Irving},
  \bibinfo{person}{Michael Isard}, \bibinfo{person}{Yangqing Jia},
  \bibinfo{person}{Rafal Jozefowicz}, \bibinfo{person}{Lukasz Kaiser},
  \bibinfo{person}{Manjunath Kudlur}, \bibinfo{person}{Josh Levenberg},
  \bibinfo{person}{Dandelion Man\'{e}}, \bibinfo{person}{Rajat Monga},
  \bibinfo{person}{Sherry Moore}, \bibinfo{person}{Derek Murray},
  \bibinfo{person}{Chris Olah}, \bibinfo{person}{Mike Schuster},
  \bibinfo{person}{Jonathon Shlens}, \bibinfo{person}{Benoit Steiner},
  \bibinfo{person}{Ilya Sutskever}, \bibinfo{person}{Kunal Talwar},
  \bibinfo{person}{Paul Tucker}, \bibinfo{person}{Vincent Vanhoucke},
  \bibinfo{person}{Vijay Vasudevan}, \bibinfo{person}{Fernanda Vi\'{e}gas},
  \bibinfo{person}{Oriol Vinyals}, \bibinfo{person}{Pete Warden},
  \bibinfo{person}{Martin Wattenberg}, \bibinfo{person}{Martin Wicke},
  \bibinfo{person}{Yuan Yu}, {and} \bibinfo{person}{Xiaoqiang Zheng}.}
  \bibinfo{year}{2015}\natexlab{}.
\newblock \bibinfo{title}{{TensorFlow}: Large-Scale Machine Learning on
  Heterogeneous Systems}.
\newblock
\newblock
\urldef\tempurl%
\url{https://www.tensorflow.org/}
\showURL{%
\tempurl}
\newblock
\shownote{Software available from tensorflow.org}.


\bibitem[Ackley et~al\mbox{.}(1985)]%
        {ackley1985learning}
\bibfield{author}{\bibinfo{person}{David~H Ackley}, \bibinfo{person}{Geoffrey~E
  Hinton}, {and} \bibinfo{person}{Terrence~J Sejnowski}.}
  \bibinfo{year}{1985}\natexlab{}.
\newblock \showarticletitle{A learning algorithm for Boltzmann machines}.
\newblock \bibinfo{journal}{\emph{Cognitive science}} \bibinfo{volume}{9},
  \bibinfo{number}{1} (\bibinfo{year}{1985}), \bibinfo{pages}{147--169}.
\newblock


\bibitem[Anderson(1999)]%
        {anderson1999point}
\bibfield{author}{\bibinfo{person}{Elizabeth~S Anderson}.}
  \bibinfo{year}{1999}\natexlab{}.
\newblock \showarticletitle{What is the Point of Equality?}
\newblock \bibinfo{journal}{\emph{Ethics}} \bibinfo{volume}{109},
  \bibinfo{number}{2} (\bibinfo{year}{1999}), \bibinfo{pages}{287--337}.
\newblock


\bibitem[Arnold et~al\mbox{.}(2019)]%
        {arnold2019factsheets}
\bibfield{author}{\bibinfo{person}{Matthew Arnold}, \bibinfo{person}{Rachel~KE
  Bellamy}, \bibinfo{person}{Michael Hind}, \bibinfo{person}{Stephanie Houde},
  \bibinfo{person}{Sameep Mehta}, \bibinfo{person}{Aleksandra Mojsilovi{\'c}},
  \bibinfo{person}{Ravi Nair}, \bibinfo{person}{K~Natesan Ramamurthy},
  \bibinfo{person}{Alexandra Olteanu}, \bibinfo{person}{David Piorkowski},
  {et~al\mbox{.}}} \bibinfo{year}{2019}\natexlab{}.
\newblock \showarticletitle{FactSheets: Increasing trust in AI services through
  supplier's declarations of conformity}.
\newblock \bibinfo{journal}{\emph{IBM Journal of Research and Development}}
  \bibinfo{volume}{63}, \bibinfo{number}{4/5} (\bibinfo{year}{2019}),
  \bibinfo{pages}{6--1}.
\newblock


\bibitem[Auer et~al\mbox{.}(2002)]%
        {auer2002finite}
\bibfield{author}{\bibinfo{person}{Peter Auer}, \bibinfo{person}{Nicolo
  Cesa-Bianchi}, {and} \bibinfo{person}{Paul Fischer}.}
  \bibinfo{year}{2002}\natexlab{}.
\newblock \showarticletitle{Finite-time analysis of the multiarmed bandit
  problem}.
\newblock \bibinfo{journal}{\emph{Machine learning}} \bibinfo{volume}{47},
  \bibinfo{number}{2} (\bibinfo{year}{2002}), \bibinfo{pages}{235--256}.
\newblock


\bibitem[Baria and Cross(2021)]%
        {baria2021brain}
\bibfield{author}{\bibinfo{person}{Alexis~T Baria} {and} \bibinfo{person}{Keith
  Cross}.} \bibinfo{year}{2021}\natexlab{}.
\newblock \showarticletitle{The brain is a computer is a brain: neuroscience's
  internal debate and the social significance of the Computational Metaphor}.
\newblock \bibinfo{journal}{\emph{arXiv preprint arXiv:2107.14042}}
  (\bibinfo{year}{2021}).
\newblock


\bibitem[Bender et~al\mbox{.}(2021)]%
        {parrots}
\bibfield{author}{\bibinfo{person}{Emily~M. Bender}, \bibinfo{person}{Timnit
  Gebru}, \bibinfo{person}{Angelina McMillan-Major}, {and}
  \bibinfo{person}{Shmargaret Shmitchell}.} \bibinfo{year}{2021}\natexlab{}.
\newblock \showarticletitle{On the Dangers of Stochastic Parrots: Can Language
  Models Be Too Big?}. In \bibinfo{booktitle}{\emph{Proceedings of the 2021 ACM
  Conference on Fairness, Accountability, and Transparency}} (Virtual Event,
  Canada) \emph{(\bibinfo{series}{FAccT '21})}. \bibinfo{publisher}{Association
  for Computing Machinery}, \bibinfo{address}{New York, NY, USA},
  \bibinfo{pages}{610–623}.
\newblock
\showISBNx{9781450383097}
\urldef\tempurl%
\url{https://doi.org/10.1145/3442188.3445922}
\showDOI{\tempurl}


\bibitem[Birhane(2021)]%
        {birhane2021impossibility}
\bibfield{author}{\bibinfo{person}{Abeba Birhane}.}
  \bibinfo{year}{2021}\natexlab{}.
\newblock \showarticletitle{The Impossibility of Automating Ambiguity}.
\newblock \bibinfo{journal}{\emph{Artificial Life}} \bibinfo{volume}{27},
  \bibinfo{number}{1} (\bibinfo{year}{2021}), \bibinfo{pages}{44--61}.
\newblock


\bibitem[Bloor(1996)]%
        {bloor1996idealism}
\bibfield{author}{\bibinfo{person}{David Bloor}.}
  \bibinfo{year}{1996}\natexlab{}.
\newblock \showarticletitle{Idealism and the Sociology of Knowledge}.
\newblock \bibinfo{journal}{\emph{Social studies of science}}
  \bibinfo{volume}{26}, \bibinfo{number}{4} (\bibinfo{year}{1996}),
  \bibinfo{pages}{839--856}.
\newblock


\bibitem[Bohannon and Charrington(2022)]%
        {twiml-nlp}
\bibfield{author}{\bibinfo{person}{John Bohannon} {and} \bibinfo{person}{Sam
  Charrington}.} \bibinfo{year}{2022}\natexlab{}.
\newblock \bibinfo{title}{Trends in NLP with John Bohannon}.
\newblock
\newblock
\urldef\tempurl%
\url{https://twimlai.com/trends-in-nlp-with-john-bohannon/}
\showURL{%
\tempurl}


\bibitem[Bommasani et~al\mbox{.}(2021)]%
        {bommasani2021opportunities}
\bibfield{author}{\bibinfo{person}{Rishi Bommasani}, \bibinfo{person}{Drew~A
  Hudson}, \bibinfo{person}{Ehsan Adeli}, \bibinfo{person}{Russ Altman},
  \bibinfo{person}{Simran Arora}, \bibinfo{person}{Sydney von Arx},
  \bibinfo{person}{Michael~S Bernstein}, \bibinfo{person}{Jeannette Bohg},
  \bibinfo{person}{Antoine Bosselut}, \bibinfo{person}{Emma Brunskill},
  {et~al\mbox{.}}} \bibinfo{year}{2021}\natexlab{}.
\newblock \showarticletitle{On the opportunities and risks of foundation
  models}.
\newblock \bibinfo{journal}{\emph{arXiv preprint arXiv:2108.07258}}
  (\bibinfo{year}{2021}).
\newblock


\bibitem[Bostrom(2017)]%
        {bostrom2017superintelligence}
\bibfield{author}{\bibinfo{person}{Nick Bostrom}.}
  \bibinfo{year}{2017}\natexlab{}.
\newblock \bibinfo{booktitle}{\emph{Superintelligence: Paths, Dangers,
  Strategies}}.
\newblock \bibinfo{publisher}{Oxford University Press}.
\newblock


\bibitem[Bostrom and Yudkowsky(2018)]%
        {bostrom2018ethics}
\bibfield{author}{\bibinfo{person}{Nick Bostrom} {and} \bibinfo{person}{Eliezer
  Yudkowsky}.} \bibinfo{year}{2018}\natexlab{}.
\newblock \showarticletitle{The ethics of artificial intelligence}.
\newblock In \bibinfo{booktitle}{\emph{Artificial intelligence safety and
  security}}, \bibfield{editor}{\bibinfo{person}{Roman~V. Yampolskiy}} (Ed.).
  \bibinfo{publisher}{Chapman and Hall/CRC}, \bibinfo{address}{Boca Raton,
  Florida}, \bibinfo{pages}{57--69}.
\newblock


\bibitem[Bowers et~al\mbox{.}(2022)]%
        {bowers2022deep}
\bibfield{author}{\bibinfo{person}{Jeffrey Bowers}, \bibinfo{person}{Gaurav
  Malhotra}, \bibinfo{person}{Marin Dujmović}, \bibinfo{person}{Milton~Llera
  Montero}, \bibinfo{person}{Christian Tsvetkov}, \bibinfo{person}{Valerio
  Biscione}, \bibinfo{person}{Guillermo Puebla}, \bibinfo{person}{Federico
  Adolfi}, \bibinfo{person}{John Hummel}, \bibinfo{person}{Rachel~Flood
  Heaton}, \bibinfo{person}{Benjamin Evans}, \bibinfo{person}{Jeff Mitchell},
  {and} \bibinfo{person}{Ryan Blything}.} \bibinfo{year}{2022}\natexlab{}.
\newblock \bibinfo{title}{Deep Problems with Neural Network Models of Human
  Vision}.
\newblock
\newblock
\urldef\tempurl%
\url{https://doi.org/10.31234/osf.io/5zf4s}
\showDOI{\tempurl}


\bibitem[Bradbury et~al\mbox{.}(2018)]%
        {jax2018github}
\bibfield{author}{\bibinfo{person}{James Bradbury}, \bibinfo{person}{Roy
  Frostig}, \bibinfo{person}{Peter Hawkins}, \bibinfo{person}{Matthew~James
  Johnson}, \bibinfo{person}{Chris Leary}, \bibinfo{person}{Dougal Maclaurin},
  \bibinfo{person}{George Necula}, \bibinfo{person}{Adam Paszke},
  \bibinfo{person}{Jake Vander{P}las}, \bibinfo{person}{Skye
  Wanderman-{M}ilne}, {and} \bibinfo{person}{Qiao Zhang}.}
  \bibinfo{year}{2018}\natexlab{}.
\newblock \bibinfo{booktitle}{\emph{{JAX}: composable transformations of
  {P}ython+{N}um{P}y programs}}.
\newblock
\urldef\tempurl%
\url{http://github.com/google/jax}
\showURL{%
\tempurl}
\newblock
\shownote{Accessed May 2, 2022}.


\bibitem[Bricken and Pehlevan(2021)]%
        {bricken2021attention}
\bibfield{author}{\bibinfo{person}{Trenton Bricken} {and}
  \bibinfo{person}{Cengiz Pehlevan}.} \bibinfo{year}{2021}\natexlab{}.
\newblock \showarticletitle{Attention Approximates Sparse Distributed Memory}.
\newblock \bibinfo{journal}{\emph{Advances in Neural Information Processing
  Systems}}  \bibinfo{volume}{34} (\bibinfo{year}{2021}).
\newblock


\bibitem[Campolo et~al\mbox{.}(2017)]%
        {ainow2017}
\bibfield{author}{\bibinfo{person}{Alex Campolo}, \bibinfo{person}{Madelyn
  Sanfilippo}, \bibinfo{person}{Meredith Whittaker}, {and}
  \bibinfo{person}{Kate Crawford}.} \bibinfo{year}{2017}\natexlab{}.
\newblock \bibinfo{title}{AI Now 2017 Report}.
\newblock
  \bibinfo{howpublished}{\url{https://ainowinstitute.org/AI_Now_2017_Report.pdf}}.
\newblock
\newblock
\shownote{Accessed: 2022-01-21}.


\bibitem[Cartwright(2021)]%
        {cartwright2021rigour}
\bibfield{author}{\bibinfo{person}{Nancy Cartwright}.}
  \bibinfo{year}{2021}\natexlab{}.
\newblock \showarticletitle{Rigour versus the need for evidential diversity}.
\newblock \bibinfo{journal}{\emph{Synthese}} \bibinfo{volume}{199},
  \bibinfo{number}{5} (\bibinfo{year}{2021}), \bibinfo{pages}{13095--13119}.
\newblock


\bibitem[Cartwright et~al\mbox{.}(1999)]%
        {cartwright1999dappled}
\bibfield{author}{\bibinfo{person}{Nancy Cartwright} {et~al\mbox{.}}}
  \bibinfo{year}{1999}\natexlab{}.
\newblock \bibinfo{booktitle}{\emph{The dappled world: A study of the
  boundaries of science}}.
\newblock \bibinfo{publisher}{Cambridge University Press}.
\newblock


\bibitem[Cat(2022)]%
        {sep-scientific-unity}
\bibfield{author}{\bibinfo{person}{Jordi Cat}.}
  \bibinfo{year}{2022}\natexlab{}.
\newblock \showarticletitle{{The Unity of Science}}.
\newblock In \bibinfo{booktitle}{\emph{The {Stanford} Encyclopedia of
  Philosophy} (\bibinfo{edition}{{S}pring 2022} ed.)},
  \bibfield{editor}{\bibinfo{person}{Edward~N. Zalta}} (Ed.).
  \bibinfo{publisher}{Metaphysics Research Lab, Stanford University}.
\newblock


\bibitem[Charrington and Gkioxari(2022)]%
        {twiml-vision}
\bibfield{author}{\bibinfo{person}{Sam Charrington} {and}
  \bibinfo{person}{Georgia Gkioxari}.} \bibinfo{year}{2022}\natexlab{}.
\newblock \bibinfo{title}{Trends in Computer Vision with Georgia Gkioxari}.
\newblock
\newblock
\urldef\tempurl%
\url{https://twimlai.com/trends-in-computer-vision-with-georgia-gkioxari/}
\showURL{%
\tempurl}


\bibitem[Chaudhari et~al\mbox{.}(2021)]%
        {chaudhari2021attentive}
\bibfield{author}{\bibinfo{person}{Sneha Chaudhari}, \bibinfo{person}{Varun
  Mithal}, \bibinfo{person}{Gungor Polatkan}, {and} \bibinfo{person}{Rohan
  Ramanath}.} \bibinfo{year}{2021}\natexlab{}.
\newblock \showarticletitle{An attentive survey of attention models}.
\newblock \bibinfo{journal}{\emph{ACM Transactions on Intelligent Systems and
  Technology (TIST)}} \bibinfo{volume}{12}, \bibinfo{number}{5}
  (\bibinfo{year}{2021}), \bibinfo{pages}{1--32}.
\newblock


\bibitem[Creel and Hellman(ming)]%
        {creel2021algorithmic}
\bibfield{author}{\bibinfo{person}{Kathleen Creel} {and}
  \bibinfo{person}{Deborah Hellman}.} \bibinfo{year}{forthcoming}\natexlab{}.
\newblock \showarticletitle{The Algorithmic Leviathan: Arbitrariness, Fairness,
  and Opportunity in Algorithmic Decision Making Systems}.
\newblock \bibinfo{journal}{\emph{Canadian Journal of Philosophy}}
  (\bibinfo{year}{forthcoming}).
\newblock


\bibitem[DeepLearning.AI(2021)]%
        {deepai2021}
\bibfield{author}{\bibinfo{person}{DeepLearning.AI}.}
  \bibinfo{year}{2021}\natexlab{}.
\newblock \bibinfo{title}{The Batch}.
\newblock
\newblock
\urldef\tempurl%
\url{https://read.deeplearning.ai/the-batch/issue-123/}
\showURL{%
\tempurl}
\newblock
\shownote{``Originally developed for natural language processing, transformers
  are becoming the Swiss Army Knife of deep learning.''}.


\bibitem[Devezer et~al\mbox{.}(2019)]%
        {devezer}
\bibfield{author}{\bibinfo{person}{Berna Devezer}, \bibinfo{person}{Luis~G.
  Nardin}, \bibinfo{person}{Bert Baumgaertner}, {and}
  \bibinfo{person}{Erkan~Ozge Buzbas}.} \bibinfo{year}{2019}\natexlab{}.
\newblock \showarticletitle{Scientific discovery in a model-centric framework:
  Reproducibility, innovation, and epistemic diversity}.
\newblock \bibinfo{journal}{\emph{PLOS ONE}} \bibinfo{volume}{14},
  \bibinfo{number}{5} (\bibinfo{date}{05} \bibinfo{year}{2019}),
  \bibinfo{pages}{1--23}.
\newblock
\urldef\tempurl%
\url{https://doi.org/10.1371/journal.pone.0216125}
\showDOI{\tempurl}


\bibitem[Dickson(2019)]%
        {moderation}
\bibfield{author}{\bibinfo{person}{Ben Dickson}.}
  \bibinfo{year}{2019}\natexlab{}.
\newblock \bibinfo{title}{Human Help Wanted: Why AI Is Terrible at Content
  Moderation}.
\newblock
\newblock
\urldef\tempurl%
\url{https://www.pcmag.com/opinions/human-help-wanted-why-ai-is-terrible-at-content-moderation}
\showURL{%
\tempurl}
\newblock
\shownote{Accessed Jan 18, 2022}.


\bibitem[Dosovitskiy et~al\mbox{.}(2020)]%
        {dosovitskiy2020image}
\bibfield{author}{\bibinfo{person}{Alexey Dosovitskiy}, \bibinfo{person}{Lucas
  Beyer}, \bibinfo{person}{Alexander Kolesnikov}, \bibinfo{person}{Dirk
  Weissenborn}, \bibinfo{person}{Xiaohua Zhai}, \bibinfo{person}{Thomas
  Unterthiner}, \bibinfo{person}{Mostafa Dehghani}, \bibinfo{person}{Matthias
  Minderer}, \bibinfo{person}{Georg Heigold}, \bibinfo{person}{Sylvain Gelly},
  {et~al\mbox{.}}} \bibinfo{year}{2020}\natexlab{}.
\newblock \showarticletitle{An image is worth 16x16 words: Transformers for
  image recognition at scale}.
\newblock \bibinfo{journal}{\emph{arXiv preprint arXiv:2010.11929}}
  (\bibinfo{year}{2020}).
\newblock


\bibitem[Ensmenger(2012)]%
        {ensmenger2012chess}
\bibfield{author}{\bibinfo{person}{Nathan Ensmenger}.}
  \bibinfo{year}{2012}\natexlab{}.
\newblock \showarticletitle{Is chess the drosophila of artificial intelligence?
  A social history of an algorithm}.
\newblock \bibinfo{journal}{\emph{Social studies of science}}
  \bibinfo{volume}{42}, \bibinfo{number}{1} (\bibinfo{year}{2012}),
  \bibinfo{pages}{5--30}.
\newblock


\bibitem[Everitt et~al\mbox{.}(2018)]%
        {everitt2018agi}
\bibfield{author}{\bibinfo{person}{Tom Everitt}, \bibinfo{person}{Gary Lea},
  {and} \bibinfo{person}{Marcus Hutter}.} \bibinfo{year}{2018}\natexlab{}.
\newblock \showarticletitle{AGI safety literature review}.
\newblock \bibinfo{journal}{\emph{arXiv preprint arXiv:1805.01109}}
  (\bibinfo{year}{2018}).
\newblock


\bibitem[Face(nd)]%
        {hugging}
\bibfield{author}{\bibinfo{person}{Hugging Face}.}
  \bibinfo{year}{n.d.}\natexlab{}.
\newblock \bibinfo{title}{huggingface (Hugging Face)}.
\newblock
\newblock
\urldef\tempurl%
\url{https://huggingface.co/huggingface}
\showURL{%
\tempurl}
\newblock
\shownote{Accessed Jan 17, 2022}.


\bibitem[Felzmann et~al\mbox{.}(2019)]%
        {felzmann2019transparency}
\bibfield{author}{\bibinfo{person}{Heike Felzmann},
  \bibinfo{person}{Eduard~Fosch Villaronga}, \bibinfo{person}{Christoph Lutz},
  {and} \bibinfo{person}{Aurelia Tam{\`o}-Larrieux}.}
  \bibinfo{year}{2019}\natexlab{}.
\newblock \showarticletitle{Transparency you can trust: Transparency
  requirements for artificial intelligence between legal norms and contextual
  concerns}.
\newblock \bibinfo{journal}{\emph{Big Data \& Society}} \bibinfo{volume}{6},
  \bibinfo{number}{1} (\bibinfo{year}{2019}),
  \bibinfo{pages}{2053951719860542}.
\newblock


\bibitem[Friedman(1974)]%
        {friedman1974explanation}
\bibfield{author}{\bibinfo{person}{Michael Friedman}.}
  \bibinfo{year}{1974}\natexlab{}.
\newblock \showarticletitle{Explanation and scientific understanding}.
\newblock \bibinfo{journal}{\emph{The Journal of Philosophy}}
  \bibinfo{volume}{71}, \bibinfo{number}{1} (\bibinfo{year}{1974}),
  \bibinfo{pages}{5--19}.
\newblock


\bibitem[Fukushima(1980)]%
        {fukushima1980self}
\bibfield{author}{\bibinfo{person}{Kunihiko Fukushima}.}
  \bibinfo{year}{1980}\natexlab{}.
\newblock \showarticletitle{A self-organizing neural network model for a
  mechanism of pattern recognition unaffected by shift in position}.
\newblock \bibinfo{journal}{\emph{Biol. Cybern.}}  \bibinfo{volume}{36}
  (\bibinfo{year}{1980}), \bibinfo{pages}{193--202}.
\newblock


\bibitem[Galison et~al\mbox{.}(1997)]%
        {galison1997image}
\bibfield{author}{\bibinfo{person}{Peter Galison} {et~al\mbox{.}}}
  \bibinfo{year}{1997}\natexlab{}.
\newblock \bibinfo{booktitle}{\emph{Image and logic: A material culture of
  microphysics}}.
\newblock \bibinfo{publisher}{University of Chicago Press}.
\newblock


\bibitem[Gebru et~al\mbox{.}(2021)]%
        {gebru2018datasheets}
\bibfield{author}{\bibinfo{person}{Timnit Gebru}, \bibinfo{person}{Jamie
  Morgenstern}, \bibinfo{person}{Briana Vecchione},
  \bibinfo{person}{Jennifer~Wortman Vaughan}, \bibinfo{person}{Hanna Wallach},
  \bibinfo{person}{Hal~Daum{\'e} Iii}, {and} \bibinfo{person}{Kate Crawford}.}
  \bibinfo{year}{2021}\natexlab{}.
\newblock \showarticletitle{Datasheets for datasets}.
\newblock \bibinfo{journal}{\emph{Commun. ACM}} \bibinfo{volume}{64},
  \bibinfo{number}{12} (\bibinfo{year}{2021}), \bibinfo{pages}{86--92}.
\newblock


\bibitem[Ghassemi et~al\mbox{.}(2021)]%
        {ghassemi2021false}
\bibfield{author}{\bibinfo{person}{Marzyeh Ghassemi}, \bibinfo{person}{Luke
  Oakden-Rayner}, {and} \bibinfo{person}{Andrew~L Beam}.}
  \bibinfo{year}{2021}\natexlab{}.
\newblock \showarticletitle{The false hope of current approaches to explainable
  artificial intelligence in health care}.
\newblock \bibinfo{journal}{\emph{The Lancet Digital Health}}
  \bibinfo{volume}{3}, \bibinfo{number}{11} (\bibinfo{year}{2021}),
  \bibinfo{pages}{e745--e750}.
\newblock


\bibitem[Ghorbani et~al\mbox{.}(2019)]%
        {ghorbani2019interpretation}
\bibfield{author}{\bibinfo{person}{Amirata Ghorbani}, \bibinfo{person}{Abubakar
  Abid}, {and} \bibinfo{person}{James Zou}.} \bibinfo{year}{2019}\natexlab{}.
\newblock \showarticletitle{Interpretation of neural networks is fragile}. In
  \bibinfo{booktitle}{\emph{Proceedings of the AAAI Conference on Artificial
  Intelligence}}, Vol.~\bibinfo{volume}{33}. \bibinfo{pages}{3681--3688}.
\newblock


\bibitem[Gilpin et~al\mbox{.}(2018)]%
        {gilpin2018explaining}
\bibfield{author}{\bibinfo{person}{Leilani~H Gilpin}, \bibinfo{person}{David
  Bau}, \bibinfo{person}{Ben~Z Yuan}, \bibinfo{person}{Ayesha Bajwa},
  \bibinfo{person}{Michael Specter}, {and} \bibinfo{person}{Lalana Kagal}.}
  \bibinfo{year}{2018}\natexlab{}.
\newblock \showarticletitle{Explaining explanations: An overview of
  interpretability of machine learning}. In \bibinfo{booktitle}{\emph{2018 IEEE
  5th International Conference on data science and advanced analytics (DSAA)}}.
  IEEE, \bibinfo{pages}{80--89}.
\newblock


\bibitem[Gitelman(2013)]%
        {gitelman2013raw}
\bibfield{editor}{\bibinfo{person}{Lisa Gitelman}} (Ed.).
  \bibinfo{year}{2013}\natexlab{}.
\newblock \bibinfo{booktitle}{\emph{Raw data is an oxymoron}}.
\newblock \bibinfo{publisher}{MIT Press}, \bibinfo{address}{Cambridge,
  Massachusetts}.
\newblock


\bibitem[Goodfellow et~al\mbox{.}(2016)]%
        {Goodfellow-et-al-2016}
\bibfield{author}{\bibinfo{person}{Ian Goodfellow}, \bibinfo{person}{Yoshua
  Bengio}, {and} \bibinfo{person}{Aaron Courville}.}
  \bibinfo{year}{2016}\natexlab{}.
\newblock \bibinfo{booktitle}{\emph{Deep Learning}}.
\newblock \bibinfo{publisher}{MIT Press}, \bibinfo{address}{Cambridge,
  Massachusetts}.
\newblock
\newblock
\shownote{\url{http://www.deeplearningbook.org}}.


\bibitem[Goyal and Bengio(2020)]%
        {goyal2020inductive}
\bibfield{author}{\bibinfo{person}{Anirudh Goyal} {and} \bibinfo{person}{Yoshua
  Bengio}.} \bibinfo{year}{2020}\natexlab{}.
\newblock \showarticletitle{Inductive biases for deep learning of higher-level
  cognition}.
\newblock \bibinfo{journal}{\emph{arXiv preprint arXiv:2011.15091}}
  (\bibinfo{year}{2020}).
\newblock


\bibitem[Grim(2019)]%
        {grim2019modeling}
\bibfield{author}{\bibinfo{person}{Patrick Grim}.}
  \bibinfo{year}{2019}\natexlab{}.
\newblock \showarticletitle{Modeling epistemology: examples and analysis in
  computational philosophy of science}. In \bibinfo{booktitle}{\emph{2019
  Spring Simulation Conference (SpringSim)}}. IEEE, \bibinfo{pages}{1--12}.
\newblock


\bibitem[Grim et~al\mbox{.}(2019)]%
        {grim2019diversity}
\bibfield{author}{\bibinfo{person}{Patrick Grim}, \bibinfo{person}{Daniel~J
  Singer}, \bibinfo{person}{Aaron Bramson}, \bibinfo{person}{Bennett Holman},
  \bibinfo{person}{Sean McGeehan}, {and} \bibinfo{person}{William~J Berger}.}
  \bibinfo{year}{2019}\natexlab{}.
\newblock \showarticletitle{Diversity, ability, and expertise in epistemic
  communities}.
\newblock \bibinfo{journal}{\emph{Philosophy of Science}} \bibinfo{volume}{86},
  \bibinfo{number}{1} (\bibinfo{year}{2019}), \bibinfo{pages}{98--123}.
\newblock


\bibitem[Haraway(1988)]%
        {haraway1988situated}
\bibfield{author}{\bibinfo{person}{Donna Haraway}.}
  \bibinfo{year}{1988}\natexlab{}.
\newblock \showarticletitle{Situated knowledges: The science question in
  feminism and the privilege of partial perspective}.
\newblock \bibinfo{journal}{\emph{Feminist studies}} \bibinfo{volume}{14},
  \bibinfo{number}{3} (\bibinfo{year}{1988}), \bibinfo{pages}{575--599}.
\newblock


\bibitem[Harding(1995)]%
        {harding1995strong}
\bibfield{author}{\bibinfo{person}{Sandra Harding}.}
  \bibinfo{year}{1995}\natexlab{}.
\newblock \showarticletitle{“Strong objectivity”: A response to the new
  objectivity question}.
\newblock \bibinfo{journal}{\emph{Synthese}} \bibinfo{volume}{104},
  \bibinfo{number}{3} (\bibinfo{year}{1995}), \bibinfo{pages}{331--349}.
\newblock


\bibitem[Heesen et~al\mbox{.}(2019)]%
        {heesen2019vindicating}
\bibfield{author}{\bibinfo{person}{Remco Heesen}, \bibinfo{person}{Liam~Kofi
  Bright}, {and} \bibinfo{person}{Andrew Zucker}.}
  \bibinfo{year}{2019}\natexlab{}.
\newblock \showarticletitle{Vindicating methodological triangulation}.
\newblock \bibinfo{journal}{\emph{Synthese}} \bibinfo{volume}{196},
  \bibinfo{number}{8} (\bibinfo{year}{2019}), \bibinfo{pages}{3067--3081}.
\newblock


\bibitem[Hissam et~al\mbox{.}(2002)]%
        {hissam2002trust}
\bibfield{author}{\bibinfo{person}{Scott~A. Hissam}, \bibinfo{person}{Daniel
  Plakosh}, {and} \bibinfo{person}{C Weinstock}.}
  \bibinfo{year}{2002}\natexlab{}.
\newblock \showarticletitle{Trust and vulnerability in open source software}.
\newblock \bibinfo{journal}{\emph{IEE Proceedings-Software}}
  \bibinfo{volume}{149}, \bibinfo{number}{1} (\bibinfo{year}{2002}),
  \bibinfo{pages}{47--51}.
\newblock


\bibitem[Hong and Page(2004)]%
        {hong2004groups}
\bibfield{author}{\bibinfo{person}{Lu Hong} {and} \bibinfo{person}{Scott~E
  Page}.} \bibinfo{year}{2004}\natexlab{}.
\newblock \showarticletitle{Groups of diverse problem solvers can outperform
  groups of high-ability problem solvers}.
\newblock \bibinfo{journal}{\emph{Proceedings of the National Academy of
  Sciences}} \bibinfo{volume}{101}, \bibinfo{number}{46}
  (\bibinfo{year}{2004}), \bibinfo{pages}{16385--16389}.
\newblock


\bibitem[Hooker(2021)]%
        {hooker2021hardware}
\bibfield{author}{\bibinfo{person}{Sara Hooker}.}
  \bibinfo{year}{2021}\natexlab{}.
\newblock \showarticletitle{The hardware lottery}.
\newblock \bibinfo{journal}{\emph{Commun. ACM}} \bibinfo{volume}{64},
  \bibinfo{number}{12} (\bibinfo{year}{2021}), \bibinfo{pages}{58--65}.
\newblock


\bibitem[Hu and Chen(2018)]%
        {hu2018short}
\bibfield{author}{\bibinfo{person}{Lily Hu} {and} \bibinfo{person}{Yiling
  Chen}.} \bibinfo{year}{2018}\natexlab{}.
\newblock \showarticletitle{A short-term intervention for long-term fairness in
  the labor market}. In \bibinfo{booktitle}{\emph{Proceedings of the 2018 World
  Wide Web Conference}}. \bibinfo{pages}{1389--1398}.
\newblock


\bibitem[Jones and Kording(2020)]%
        {jones2020can}
\bibfield{author}{\bibinfo{person}{Ilenna~Simone Jones} {and}
  \bibinfo{person}{Konrad~Paul Kording}.} \bibinfo{year}{2020}\natexlab{}.
\newblock \showarticletitle{Can single neurons solve {MNIST}? The computational
  power of biological dendritic trees}.
\newblock \bibinfo{journal}{\emph{arXiv preprint arXiv:2009.01269}}
  (\bibinfo{year}{2020}).
\newblock


\bibitem[@karpathy (Andrej~Karpathy)(2021)]%
        {karpathy}
\bibfield{author}{\bibinfo{person}{@karpathy (Andrej~Karpathy)}.}
  \bibinfo{year}{2021}\natexlab{}.
\newblock \bibinfo{title}{The ongoing consolidation in AI is incredible....}
\newblock
\newblock
\urldef\tempurl%
\url{https://twitter.com/karpathy/status/1468370605229547522?s=20&t=go3X-8IlBf_X-ekQ07oh7g}
\showURL{%
\tempurl}
\newblock
\shownote{Accessed: 2022-05-01}.


\bibitem[Kelly et~al\mbox{.}(2019)]%
        {kelly2019key}
\bibfield{author}{\bibinfo{person}{Christopher~J Kelly}, \bibinfo{person}{Alan
  Karthikesalingam}, \bibinfo{person}{Mustafa Suleyman}, \bibinfo{person}{Greg
  Corrado}, {and} \bibinfo{person}{Dominic King}.}
  \bibinfo{year}{2019}\natexlab{}.
\newblock \showarticletitle{Key challenges for delivering clinical impact with
  artificial intelligence}.
\newblock \bibinfo{journal}{\emph{BMC Medicine}} \bibinfo{volume}{17},
  \bibinfo{number}{1} (\bibinfo{year}{2019}), \bibinfo{pages}{1--9}.
\newblock


\bibitem[Kitcher(1981)]%
        {kitcher1981explanatory}
\bibfield{author}{\bibinfo{person}{Philip Kitcher}.}
  \bibinfo{year}{1981}\natexlab{}.
\newblock \showarticletitle{Explanatory unification}.
\newblock \bibinfo{journal}{\emph{Philosophy of Science}} \bibinfo{volume}{48},
  \bibinfo{number}{4} (\bibinfo{year}{1981}), \bibinfo{pages}{507--531}.
\newblock


\bibitem[Kleinberg and Raghavan(2021)]%
        {Kleinberge2018340118}
\bibfield{author}{\bibinfo{person}{Jon Kleinberg} {and} \bibinfo{person}{Manish
  Raghavan}.} \bibinfo{year}{2021}\natexlab{}.
\newblock \showarticletitle{Algorithmic monoculture and social welfare}.
\newblock \bibinfo{journal}{\emph{Proceedings of the National Academy of
  Sciences}} \bibinfo{volume}{118}, \bibinfo{number}{22}
  (\bibinfo{year}{2021}).
\newblock
\showISSN{0027-8424}
\urldef\tempurl%
\url{https://doi.org/10.1073/pnas.2018340118}
\showDOI{\tempurl}


\bibitem[Kockelman(2020)]%
        {kockelman2020epistemic}
\bibfield{author}{\bibinfo{person}{Paul Kockelman}.}
  \bibinfo{year}{2020}\natexlab{}.
\newblock \showarticletitle{The epistemic and performative dynamics of machine
  learning praxis}.
\newblock \bibinfo{journal}{\emph{Signs and Society}} \bibinfo{volume}{8},
  \bibinfo{number}{2} (\bibinfo{year}{2020}), \bibinfo{pages}{319--355}.
\newblock


\bibitem[Lake et~al\mbox{.}(2017)]%
        {lake2017building}
\bibfield{author}{\bibinfo{person}{Brenden~M Lake}, \bibinfo{person}{Tomer~D
  Ullman}, \bibinfo{person}{Joshua~B Tenenbaum}, {and}
  \bibinfo{person}{Samuel~J Gershman}.} \bibinfo{year}{2017}\natexlab{}.
\newblock \showarticletitle{Building machines that learn and think like
  people}.
\newblock \bibinfo{journal}{\emph{Behavioral and Brain Sciences}}
  \bibinfo{volume}{40} (\bibinfo{year}{2017}).
\newblock


\bibitem[Lazzaro(2021)]%
        {diverse}
\bibfield{author}{\bibinfo{person}{Sage Lazzaro}.}
  \bibinfo{year}{2021}\natexlab{}.
\newblock \bibinfo{title}{Are AI ethics teams doomed to be a facade? Women who
  pioneered them weigh in}.
\newblock
\newblock
\urldef\tempurl%
\url{https://venturebeat.com/2021/09/30/are-ai-ethics-teams-doomed-to-be-a-facade-the-women-who-pioneered-them-weigh-in/}
\showURL{%
\tempurl}
\newblock
\shownote{Accessed Jan 18, 2022}.


\bibitem[Lee et~al\mbox{.}(2015)]%
        {lee2015m}
\bibfield{author}{\bibinfo{person}{Stefan Lee}, \bibinfo{person}{Senthil
  Purushwalkam}, \bibinfo{person}{Michael Cogswell}, \bibinfo{person}{David
  Crandall}, {and} \bibinfo{person}{Dhruv Batra}.}
  \bibinfo{year}{2015}\natexlab{}.
\newblock \showarticletitle{Why m heads are better than one: Training a diverse
  ensemble of deep networks}.
\newblock \bibinfo{journal}{\emph{arXiv preprint arXiv:1511.06314}}
  (\bibinfo{year}{2015}).
\newblock


\bibitem[Liebowitz and Margolis(1995)]%
        {liebowitz1995path}
\bibfield{author}{\bibinfo{person}{Stan~J Liebowitz} {and}
  \bibinfo{person}{Stephen~E Margolis}.} \bibinfo{year}{1995}\natexlab{}.
\newblock \showarticletitle{Path dependence, lock-in, and history}.
\newblock \bibinfo{journal}{\emph{Journal of Law, Economics, \& Organization}}
  \bibinfo{volume}{11} (\bibinfo{year}{1995}), \bibinfo{pages}{205--226}.
\newblock
Issue 1.


\bibitem[Liu et~al\mbox{.}(2022)]%
        {liu2022convnet}
\bibfield{author}{\bibinfo{person}{Zhuang Liu}, \bibinfo{person}{Hanzi Mao},
  \bibinfo{person}{Chao-Yuan Wu}, \bibinfo{person}{Christoph Feichtenhofer},
  \bibinfo{person}{Trevor Darrell}, {and} \bibinfo{person}{Saining Xie}.}
  \bibinfo{year}{2022}\natexlab{}.
\newblock \showarticletitle{A ConvNet for the 2020s}.
\newblock \bibinfo{journal}{\emph{arXiv preprint arXiv:2201.03545}}
  (\bibinfo{year}{2022}).
\newblock


\bibitem[Mahoney(2000)]%
        {mahoney2000path}
\bibfield{author}{\bibinfo{person}{James Mahoney}.}
  \bibinfo{year}{2000}\natexlab{}.
\newblock \showarticletitle{Path dependence in historical sociology}.
\newblock \bibinfo{journal}{\emph{Theory and Society}} \bibinfo{volume}{29},
  \bibinfo{number}{4} (\bibinfo{year}{2000}), \bibinfo{pages}{507--548}.
\newblock


\bibitem[M{\"a}ki(2001)]%
        {maki2001explanatory}
\bibfield{author}{\bibinfo{person}{Uskali M{\"a}ki}.}
  \bibinfo{year}{2001}\natexlab{}.
\newblock \showarticletitle{Explanatory unification: Double and doubtful}.
\newblock \bibinfo{journal}{\emph{Philosophy of the Social Sciences}}
  \bibinfo{volume}{31}, \bibinfo{number}{4} (\bibinfo{year}{2001}),
  \bibinfo{pages}{488--506}.
\newblock


\bibitem[McClelland et~al\mbox{.}(2020)]%
        {mcclelland2020placing}
\bibfield{author}{\bibinfo{person}{James~L McClelland}, \bibinfo{person}{Felix
  Hill}, \bibinfo{person}{Maja Rudolph}, \bibinfo{person}{Jason Baldridge},
  {and} \bibinfo{person}{Hinrich Sch{\"u}tze}.}
  \bibinfo{year}{2020}\natexlab{}.
\newblock \showarticletitle{Placing language in an integrated understanding
  system: Next steps toward human-level performance in neural language models}.
\newblock \bibinfo{journal}{\emph{Proceedings of the National Academy of
  Sciences}} \bibinfo{volume}{117}, \bibinfo{number}{42}
  (\bibinfo{year}{2020}), \bibinfo{pages}{25966--25974}.
\newblock


\bibitem[McClelland et~al\mbox{.}(1986)]%
        {mcclelland1986parallel}
\bibfield{author}{\bibinfo{person}{James~L McClelland},
  \bibinfo{person}{David~E Rumelhart}, \bibinfo{person}{PDP~Research Group},
  {et~al\mbox{.}}} \bibinfo{year}{1986}\natexlab{}.
\newblock \bibinfo{booktitle}{\emph{Parallel distributed processing}}.
  Vol.~\bibinfo{volume}{2}.
\newblock \bibinfo{publisher}{MIT Press}, \bibinfo{address}{Cambridge,
  Massachusetts}.
\newblock


\bibitem[McCulloch and Pitts(1990)]%
        {mcculloch1990logical}
\bibfield{author}{\bibinfo{person}{Warren~S McCulloch} {and}
  \bibinfo{person}{Walter Pitts}.} \bibinfo{year}{1990}\natexlab{}.
\newblock \showarticletitle{A logical calculus of the ideas immanent in nervous
  activity}.
\newblock \bibinfo{journal}{\emph{Bulletin of mathematical biology}}
  \bibinfo{volume}{52}, \bibinfo{number}{1} (\bibinfo{year}{1990}),
  \bibinfo{pages}{99--115}.
\newblock


\bibitem[McLennan et~al\mbox{.}(2020)]%
        {mclennan2020ai}
\bibfield{author}{\bibinfo{person}{Stuart McLennan},
  \bibinfo{person}{Meredith~M Lee}, \bibinfo{person}{Amelia Fiske}, {and}
  \bibinfo{person}{Leo~Anthony Celi}.} \bibinfo{year}{2020}\natexlab{}.
\newblock \showarticletitle{AI ethics is not a panacea}.
\newblock \bibinfo{journal}{\emph{The American Journal of Bioethics}}
  \bibinfo{volume}{20}, \bibinfo{number}{11} (\bibinfo{year}{2020}),
  \bibinfo{pages}{20--22}.
\newblock


\bibitem[Melanson(2021)]%
        {log4j}
\bibfield{author}{\bibinfo{person}{Mike Melanson}.}
  \bibinfo{year}{2021}\natexlab{}.
\newblock \bibinfo{title}{Log4j Is One Big ``I Told You So'' for Open Source
  Communities}.
\newblock
\newblock
\urldef\tempurl%
\url{https://thenewstack.io/log4j-is-one-big-i-told-you-so-for-open-source-communities/}
\showURL{%
\tempurl}
\newblock
\shownote{Accessed Jan 17, 2022}.


\bibitem[Mellow~Jr.(2022)]%
        {scale2022}
\bibfield{author}{\bibinfo{person}{John.~P. Mellow~Jr.}}
  \bibinfo{year}{2022}\natexlab{}.
\newblock \bibinfo{title}{State of AI Report: Transformers Are Taking the AI
  World by Storm}.
\newblock
\newblock
\urldef\tempurl%
\url{https://exchange.scale.com/public/blogs/state-of-ai-report-2021-transformers-taking-ai-world-by-storm-nathan-benaich}
\showURL{%
\tempurl}
\newblock
\shownote{``Transformers [...] took the machine learning world by storm in
  2021. Originally designed to work with natural language processing models,
  the technology has blasted out of NLP over the last 12 months to emerge as a
  general-purpose architecture for ML.'' Accessed May 9, 2022}.


\bibitem[Melville and Mooney(2004)]%
        {melville2004diverse}
\bibfield{author}{\bibinfo{person}{Prem Melville} {and}
  \bibinfo{person}{Raymond~J Mooney}.} \bibinfo{year}{2004}\natexlab{}.
\newblock \showarticletitle{Diverse ensembles for active learning}. In
  \bibinfo{booktitle}{\emph{Proceedings of the twenty-first international
  conference on Machine learning}}. \bibinfo{pages}{74}.
\newblock


\bibitem[Minsky and Papert(2017)]%
        {minsky2017perceptrons}
\bibfield{author}{\bibinfo{person}{Marvin Minsky} {and}
  \bibinfo{person}{Seymour~A Papert}.} \bibinfo{year}{2017}\natexlab{}.
\newblock \bibinfo{booktitle}{\emph{Perceptrons: An introduction to
  computational geometry}}.
\newblock \bibinfo{publisher}{MIT Press}, \bibinfo{address}{Cambridge,
  Massachusetts}.
\newblock


\bibitem[Mitchell(2021)]%
        {mitchell2021ai}
\bibfield{author}{\bibinfo{person}{Melanie Mitchell}.}
  \bibinfo{year}{2021}\natexlab{}.
\newblock \showarticletitle{Why AI is harder than we think}.
\newblock \bibinfo{journal}{\emph{arXiv preprint arXiv:2104.12871}}
  (\bibinfo{year}{2021}).
\newblock


\bibitem[M{\"u}ller and Bostrom(2016)]%
        {muller2016future}
\bibfield{author}{\bibinfo{person}{Vincent~C M{\"u}ller} {and}
  \bibinfo{person}{Nick Bostrom}.} \bibinfo{year}{2016}\natexlab{}.
\newblock \showarticletitle{Future progress in artificial intelligence: A
  survey of expert opinion}.
\newblock In \bibinfo{booktitle}{\emph{Fundamental issues of artificial
  intelligence}}. \bibinfo{publisher}{Springer}, \bibinfo{pages}{555--572}.
\newblock


\bibitem[Musgrave et~al\mbox{.}(2020)]%
        {musgrave2020metric}
\bibfield{author}{\bibinfo{person}{Kevin Musgrave}, \bibinfo{person}{Serge
  Belongie}, {and} \bibinfo{person}{Ser-Nam Lim}.}
  \bibinfo{year}{2020}\natexlab{}.
\newblock \showarticletitle{A metric learning reality check}. In
  \bibinfo{booktitle}{\emph{European Conference on Computer Vision}}. Springer,
  \bibinfo{pages}{681--699}.
\newblock


\bibitem[Myrvold(2003)]%
        {myrvold2003bayesian}
\bibfield{author}{\bibinfo{person}{Wayne~C Myrvold}.}
  \bibinfo{year}{2003}\natexlab{}.
\newblock \showarticletitle{A Bayesian account of the virtue of unification}.
\newblock \bibinfo{journal}{\emph{Philosophy of Science}} \bibinfo{volume}{70},
  \bibinfo{number}{2} (\bibinfo{year}{2003}), \bibinfo{pages}{399--423}.
\newblock


\bibitem[Nagel(1989)]%
        {nagel1989view}
\bibfield{author}{\bibinfo{person}{Thomas Nagel}.}
  \bibinfo{year}{1989}\natexlab{}.
\newblock \bibinfo{booktitle}{\emph{The view from nowhere}}.
\newblock \bibinfo{publisher}{Oxford University Press}.
\newblock


\bibitem[Neyshabur et~al\mbox{.}(2014)]%
        {neyshabur2014search}
\bibfield{author}{\bibinfo{person}{Behnam Neyshabur}, \bibinfo{person}{Ryota
  Tomioka}, {and} \bibinfo{person}{Nathan Srebro}.}
  \bibinfo{year}{2014}\natexlab{}.
\newblock \showarticletitle{In search of the real inductive bias: On the role
  of implicit regularization in deep learning}.
\newblock \bibinfo{journal}{\emph{arXiv preprint arXiv:1412.6614}}
  (\bibinfo{year}{2014}).
\newblock


\bibitem[Noble(2018)]%
        {noble2018algorithms}
\bibfield{author}{\bibinfo{person}{Safiya~Umoja Noble}.}
  \bibinfo{year}{2018}\natexlab{}.
\newblock \bibinfo{booktitle}{\emph{Algorithms of oppression}}.
\newblock \bibinfo{publisher}{New York University Press}.
\newblock


\bibitem[Olazaran(1996)]%
        {olazaran1996sociological}
\bibfield{author}{\bibinfo{person}{Mikel Olazaran}.}
  \bibinfo{year}{1996}\natexlab{}.
\newblock \showarticletitle{A sociological study of the official history of the
  perceptrons controversy}.
\newblock \bibinfo{journal}{\emph{Social Studies of Science}}
  \bibinfo{volume}{26}, \bibinfo{number}{3} (\bibinfo{year}{1996}),
  \bibinfo{pages}{611--659}.
\newblock


\bibitem[O'Reilly(1998)]%
        {o1998six}
\bibfield{author}{\bibinfo{person}{Randall~C O'Reilly}.}
  \bibinfo{year}{1998}\natexlab{}.
\newblock \showarticletitle{Six principles for biologically based computational
  models of cortical cognition}.
\newblock \bibinfo{journal}{\emph{Trends in cognitive sciences}}
  \bibinfo{volume}{2}, \bibinfo{number}{11} (\bibinfo{year}{1998}),
  \bibinfo{pages}{455--462}.
\newblock


\bibitem[Ornes(2022)]%
        {ornes2022}
\bibfield{author}{\bibinfo{person}{Stephen Ornes}.}
  \bibinfo{year}{2022}\natexlab{}.
\newblock \bibinfo{title}{Will Transformers Take Over Artificial Intelligence?}
\newblock
\newblock
\urldef\tempurl%
\url{https://www.quantamagazine.org/will-transformers-take-over-artificial-intelligence-20220310/}
\showURL{%
\tempurl}
\newblock
\shownote{``Wang, for example, thinks the transformer may be a big step toward
  achieving a kind of convergence of neural net architectures, resulting in a
  universal approach to computer vision — and perhaps to other AI tasks as
  well.'' Accessed May 9, 2022}.


\bibitem[{Papers With Code}(2022a)]%
        {sota2022imagenet}
\bibfield{author}{\bibinfo{person}{{Papers With Code}}.}
  \bibinfo{year}{2022}\natexlab{a}.
\newblock \bibinfo{title}{Image Classification on ImageNet}.
\newblock
\newblock
\urldef\tempurl%
\url{https://paperswithcode.com/sota/image-classification-on-imagenet}
\showURL{%
\tempurl}
\newblock
\shownote{Accessed Apr 22, 2022}.


\bibitem[{Papers With Code}(2022b)]%
        {sota2022translation}
\bibfield{author}{\bibinfo{person}{{Papers With Code}}.}
  \bibinfo{year}{2022}\natexlab{b}.
\newblock \bibinfo{title}{Machine Translation on WMT2014 English-German}.
\newblock
\newblock
\urldef\tempurl%
\url{https://paperswithcode.com/sota/machine-translation-on-wmt2014-english-german}
\showURL{%
\tempurl}
\newblock
\shownote{Accessed Apr 22, 2022}.


\bibitem[{Papers With Code}(2022c)]%
        {sota2022object}
\bibfield{author}{\bibinfo{person}{{Papers With Code}}.}
  \bibinfo{year}{2022}\natexlab{c}.
\newblock \bibinfo{title}{Object Detection on COCO test-dev}.
\newblock
\newblock
\urldef\tempurl%
\url{https://paperswithcode.com/sota/object-detection-on-coco}
\showURL{%
\tempurl}
\newblock
\shownote{Accessed Apr 22, 2022}.


\bibitem[{Papers With Code}(2022d)]%
        {sota2022qa}
\bibfield{author}{\bibinfo{person}{{Papers With Code}}.}
  \bibinfo{year}{2022}\natexlab{d}.
\newblock \bibinfo{title}{Question Answering on SQuAD1.1}.
\newblock
\newblock
\urldef\tempurl%
\url{https://paperswithcode.com/sota/question-answering-on-squad11}
\showURL{%
\tempurl}
\newblock
\shownote{Accessed Apr 22, 2022}.


\bibitem[{Papers With Code}(2022e)]%
        {sota2022segment}
\bibfield{author}{\bibinfo{person}{{Papers With Code}}.}
  \bibinfo{year}{2022}\natexlab{e}.
\newblock \bibinfo{title}{Semantic Segmentation on ADE20K}.
\newblock
\newblock
\urldef\tempurl%
\url{https://paperswithcode.com/sota/semantic-segmentation-on-ade20k}
\showURL{%
\tempurl}
\newblock
\shownote{Accessed Apr 22, 2022}.


\bibitem[{Papers With Code}(2022f)]%
        {sota2022sentiment}
\bibfield{author}{\bibinfo{person}{{Papers With Code}}.}
  \bibinfo{year}{2022}\natexlab{f}.
\newblock \bibinfo{title}{Sentiment Analysis on SST-2 Binary classification}.
\newblock
\newblock
\urldef\tempurl%
\url{https://paperswithcode.com/sota/sentiment-analysis-on-sst-2-binary}
\showURL{%
\tempurl}
\newblock
\shownote{Accessed Apr 22, 2022}.


\bibitem[{Papers With Code}(2022g)]%
        {sota2022speech}
\bibfield{author}{\bibinfo{person}{{Papers With Code}}.}
  \bibinfo{year}{2022}\natexlab{g}.
\newblock \bibinfo{title}{Speech Recognition on LibriSpeech test-clean}.
\newblock
\newblock
\urldef\tempurl%
\url{https://paperswithcode.com/sota/speech-recognition-on-librispeech-test-clean}
\showURL{%
\tempurl}
\newblock
\shownote{Accessed Apr 22, 2022}.


\bibitem[Papert(1988)]%
        {papert1988one}
\bibfield{author}{\bibinfo{person}{Seymour Papert}.}
  \bibinfo{year}{1988}\natexlab{}.
\newblock \showarticletitle{One AI or many?}
\newblock \bibinfo{journal}{\emph{Daedalus}}  \bibinfo{volume}{117}
  (\bibinfo{year}{1988}), \bibinfo{pages}{1--14}.
\newblock
Issue 1.


\bibitem[Paszke et~al\mbox{.}(2019)]%
        {NEURIPS2019_9015}
\bibfield{author}{\bibinfo{person}{Adam Paszke}, \bibinfo{person}{Sam Gross},
  \bibinfo{person}{Francisco Massa}, \bibinfo{person}{Adam Lerer},
  \bibinfo{person}{James Bradbury}, \bibinfo{person}{Gregory Chanan},
  \bibinfo{person}{Trevor Killeen}, \bibinfo{person}{Zeming Lin},
  \bibinfo{person}{Natalia Gimelshein}, \bibinfo{person}{Luca Antiga},
  \bibinfo{person}{Alban Desmaison}, \bibinfo{person}{Andreas Kopf},
  \bibinfo{person}{Edward Yang}, \bibinfo{person}{Zachary DeVito},
  \bibinfo{person}{Martin Raison}, \bibinfo{person}{Alykhan Tejani},
  \bibinfo{person}{Sasank Chilamkurthy}, \bibinfo{person}{Benoit Steiner},
  \bibinfo{person}{Lu Fang}, \bibinfo{person}{Junjie Bai}, {and}
  \bibinfo{person}{Soumith Chintala}.} \bibinfo{year}{2019}\natexlab{}.
\newblock \showarticletitle{PyTorch: An Imperative Style, High-Performance Deep
  Learning Library}.
\newblock In \bibinfo{booktitle}{\emph{Advances in Neural Information
  Processing Systems 32}}, \bibfield{editor}{\bibinfo{person}{H.~Wallach},
  \bibinfo{person}{H.~Larochelle}, \bibinfo{person}{A.~Beygelzimer},
  \bibinfo{person}{F.~d\textquotesingle Alch\'{e}-Buc},
  \bibinfo{person}{E.~Fox}, {and} \bibinfo{person}{R.~Garnett}} (Eds.).
  \bibinfo{publisher}{Curran Associates, Inc.}, \bibinfo{pages}{8024--8035}.
\newblock
\urldef\tempurl%
\url{http://papers.neurips.cc/paper/9015-pytorch-an-imperative-style-high-performance-deep-learning-library.pdf}
\showURL{%
\tempurl}


\bibitem[Plutynski(2005)]%
        {plutynski2005explanatory}
\bibfield{author}{\bibinfo{person}{Anya Plutynski}.}
  \bibinfo{year}{2005}\natexlab{}.
\newblock \showarticletitle{Explanatory unification and the early synthesis}.
\newblock \bibinfo{journal}{\emph{The British journal for the philosophy of
  science}} \bibinfo{volume}{56}, \bibinfo{number}{3} (\bibinfo{year}{2005}),
  \bibinfo{pages}{595--609}.
\newblock


\bibitem[Prakash et~al\mbox{.}(2021)]%
        {prakash2021multi}
\bibfield{author}{\bibinfo{person}{Aditya Prakash}, \bibinfo{person}{Kashyap
  Chitta}, {and} \bibinfo{person}{Andreas Geiger}.}
  \bibinfo{year}{2021}\natexlab{}.
\newblock \showarticletitle{Multi-modal fusion transformer for end-to-end
  autonomous driving}. In \bibinfo{booktitle}{\emph{Proceedings of the IEEE/CVF
  Conference on Computer Vision and Pattern Recognition}}.
  \bibinfo{pages}{7077--7087}.
\newblock


\bibitem[Raji et~al\mbox{.}(2021)]%
        {raji2021ai}
\bibfield{author}{\bibinfo{person}{Inioluwa~Deborah Raji},
  \bibinfo{person}{Emily~M Bender}, \bibinfo{person}{Amandalynne Paullada},
  \bibinfo{person}{Emily Denton}, {and} \bibinfo{person}{Alex Hanna}.}
  \bibinfo{year}{2021}\natexlab{}.
\newblock \showarticletitle{AI and the everything in the whole wide world
  benchmark}.
\newblock \bibinfo{journal}{\emph{Conference on Neural Information Processing
  Systems (NeurIPS 2021) Track on Datasets and Benchmarks}}
  (\bibinfo{year}{2021}).
\newblock


\bibitem[Rao et~al\mbox{.}(2020)]%
        {rao2020transformer}
\bibfield{author}{\bibinfo{person}{Roshan Rao}, \bibinfo{person}{Joshua Meier},
  \bibinfo{person}{Tom Sercu}, \bibinfo{person}{Sergey Ovchinnikov}, {and}
  \bibinfo{person}{Alexander Rives}.} \bibinfo{year}{2020}\natexlab{}.
\newblock \showarticletitle{Transformer protein language models are
  unsupervised structure learners}. In \bibinfo{booktitle}{\emph{International
  Conference on Learning Representations}}.
\newblock


\bibitem[Raz(1986)]%
        {raz1986morality}
\bibfield{author}{\bibinfo{person}{Joseph Raz}.}
  \bibinfo{year}{1986}\natexlab{}.
\newblock \bibinfo{booktitle}{\emph{The morality of freedom}}.
\newblock \bibinfo{publisher}{Clarendon Press}, \bibinfo{address}{Oxford}.
\newblock


\bibitem[Research(2021)]%
        {msft2022}
\bibfield{author}{\bibinfo{person}{Microsoft Research}.}
  \bibinfo{year}{2021}\natexlab{}.
\newblock \bibinfo{title}{Five reasons to embrace Transformer in computer
  vision}.
\newblock
\newblock
\urldef\tempurl%
\url{https://www.microsoft.com/en-us/research/lab/microsoft-research-asia/articles/five-reasons-to-embrace-transformer-in-computer-vision/}
\showURL{%
\tempurl}
\newblock
\shownote{``Unity is a common goal in many disciplines. [...] In natural
  language processing (NLP), the current dominant modeling network is
  Transformer; in computer vision (CV), for a long time, the dominant
  architecture was convolutional neural networks (CNN); in social computing
  field, the dominant architecture is graph networks; and so on. However, the
  situation has changed since the end of last year, when Transformers began to
  demonstrate revolutionary performance improvements for a variety of computer
  vision tasks.'' Accessed May 9, 2022}.


\bibitem[Richards et~al\mbox{.}(2019)]%
        {richards2019deep}
\bibfield{author}{\bibinfo{person}{Blake~A Richards},
  \bibinfo{person}{Timothy~P Lillicrap}, \bibinfo{person}{Philippe Beaudoin},
  \bibinfo{person}{Yoshua Bengio}, \bibinfo{person}{Rafal Bogacz},
  \bibinfo{person}{Amelia Christensen}, \bibinfo{person}{Claudia Clopath},
  \bibinfo{person}{Rui~Ponte Costa}, \bibinfo{person}{Archy de Berker},
  \bibinfo{person}{Surya Ganguli}, {et~al\mbox{.}}}
  \bibinfo{year}{2019}\natexlab{}.
\newblock \showarticletitle{A deep learning framework for neuroscience}.
\newblock \bibinfo{journal}{\emph{Nature neuroscience}} \bibinfo{volume}{22},
  \bibinfo{number}{11} (\bibinfo{year}{2019}), \bibinfo{pages}{1761--1770}.
\newblock


\bibitem[Robeyns and Byskov(2021)]%
        {sep-capability-approach}
\bibfield{author}{\bibinfo{person}{Ingrid Robeyns} {and}
  \bibinfo{person}{Morten~Fibieger Byskov}.} \bibinfo{year}{2021}\natexlab{}.
\newblock \showarticletitle{{The Capability Approach}}.
\newblock In \bibinfo{booktitle}{\emph{The {Stanford} Encyclopedia of
  Philosophy} (\bibinfo{edition}{{W}inter 2021} ed.)},
  \bibfield{editor}{\bibinfo{person}{Edward~N. Zalta}} (Ed.).
  \bibinfo{publisher}{Metaphysics Research Lab, Stanford University}.
\newblock


\bibitem[Rosenblatt(1961)]%
        {rosenblatt1961principles}
\bibfield{author}{\bibinfo{person}{Frank Rosenblatt}.}
  \bibinfo{year}{1961}\natexlab{}.
\newblock \bibinfo{booktitle}{\emph{Principles of neurodynamics. perceptrons
  and the theory of brain mechanisms}}.
\newblock \bibinfo{type}{{T}echnical {R}eport}. \bibinfo{institution}{Cornell
  Aeronautical Lab Inc Buffalo NY}.
\newblock


\bibitem[Roy et~al\mbox{.}(2019)]%
        {roy2019towards}
\bibfield{author}{\bibinfo{person}{Kaushik Roy}, \bibinfo{person}{Akhilesh
  Jaiswal}, {and} \bibinfo{person}{Priyadarshini Panda}.}
  \bibinfo{year}{2019}\natexlab{}.
\newblock \showarticletitle{Towards spike-based machine intelligence with
  neuromorphic computing}.
\newblock \bibinfo{journal}{\emph{Nature}} \bibinfo{volume}{575},
  \bibinfo{number}{7784} (\bibinfo{year}{2019}), \bibinfo{pages}{607--617}.
\newblock


\bibitem[Rumelhart et~al\mbox{.}(1986)]%
        {rumelhart1986learning}
\bibfield{author}{\bibinfo{person}{David~E Rumelhart},
  \bibinfo{person}{Geoffrey~E Hinton}, {and} \bibinfo{person}{Ronald~J
  Williams}.} \bibinfo{year}{1986}\natexlab{}.
\newblock \showarticletitle{Learning representations by back-propagating
  errors}.
\newblock \bibinfo{journal}{\emph{Nature}} \bibinfo{volume}{323},
  \bibinfo{number}{6088} (\bibinfo{year}{1986}), \bibinfo{pages}{533--536}.
\newblock


\bibitem[Schrittwieser et~al\mbox{.}(2020)]%
        {schrittwieser2020mastering}
\bibfield{author}{\bibinfo{person}{Julian Schrittwieser},
  \bibinfo{person}{Ioannis Antonoglou}, \bibinfo{person}{Thomas Hubert},
  \bibinfo{person}{Karen Simonyan}, \bibinfo{person}{Laurent Sifre},
  \bibinfo{person}{Simon Schmitt}, \bibinfo{person}{Arthur Guez},
  \bibinfo{person}{Edward Lockhart}, \bibinfo{person}{Demis Hassabis},
  \bibinfo{person}{Thore Graepel}, {et~al\mbox{.}}}
  \bibinfo{year}{2020}\natexlab{}.
\newblock \showarticletitle{Mastering Atari, Go, chess and shogi by planning
  with a learned model}.
\newblock \bibinfo{journal}{\emph{Nature}} \bibinfo{volume}{588},
  \bibinfo{number}{7839} (\bibinfo{year}{2020}), \bibinfo{pages}{604--609}.
\newblock


\bibitem[Sejnowski(2020)]%
        {sejnowski2020unreasonable}
\bibfield{author}{\bibinfo{person}{Terrence~J Sejnowski}.}
  \bibinfo{year}{2020}\natexlab{}.
\newblock \showarticletitle{The unreasonable effectiveness of deep learning in
  artificial intelligence}.
\newblock \bibinfo{journal}{\emph{Proceedings of the National Academy of
  Sciences}} \bibinfo{volume}{117}, \bibinfo{number}{48}
  (\bibinfo{year}{2020}), \bibinfo{pages}{30033--30038}.
\newblock


\bibitem[Selbst et~al\mbox{.}(2019)]%
        {portability}
\bibfield{author}{\bibinfo{person}{Andrew~D. Selbst}, \bibinfo{person}{Danah
  Boyd}, \bibinfo{person}{Sorelle~A. Friedler}, \bibinfo{person}{Suresh
  Venkatasubramanian}, {and} \bibinfo{person}{Janet Vertesi}.}
  \bibinfo{year}{2019}\natexlab{}.
\newblock \showarticletitle{Fairness and Abstraction in Sociotechnical
  Systems}. In \bibinfo{booktitle}{\emph{Proceedings of the Conference on
  Fairness, Accountability, and Transparency}} (Atlanta, GA, USA)
  \emph{(\bibinfo{series}{FAT* '19})}. \bibinfo{publisher}{Association for
  Computing Machinery}, \bibinfo{address}{New York, NY, USA},
  \bibinfo{pages}{59–68}.
\newblock
\showISBNx{9781450361255}
\urldef\tempurl%
\url{https://doi.org/10.1145/3287560.3287598}
\showDOI{\tempurl}


\bibitem[Shevlin et~al\mbox{.}(2019)]%
        {shevlin2019limits}
\bibfield{author}{\bibinfo{person}{Henry Shevlin}, \bibinfo{person}{Karina
  Vold}, \bibinfo{person}{Matthew Crosby}, {and} \bibinfo{person}{Marta
  Halina}.} \bibinfo{year}{2019}\natexlab{}.
\newblock \showarticletitle{The limits of machine intelligence: Despite
  progress in machine intelligence, artificial general intelligence is still a
  major challenge}.
\newblock \bibinfo{journal}{\emph{EMBO Reports}} \bibinfo{volume}{20},
  \bibinfo{number}{10} (\bibinfo{year}{2019}), \bibinfo{pages}{e49177}.
\newblock


\bibitem[Skipper~Jr(1999)]%
        {skipper1999selection}
\bibfield{author}{\bibinfo{person}{Robert~A Skipper~Jr}.}
  \bibinfo{year}{1999}\natexlab{}.
\newblock \showarticletitle{Selection and the extent of explanatory
  unification}.
\newblock \bibinfo{journal}{\emph{Philosophy of Science}}  \bibinfo{volume}{66}
  (\bibinfo{year}{1999}), \bibinfo{pages}{S196--S209}.
\newblock


\bibitem[Sober(2003)]%
        {sober2003two}
\bibfield{author}{\bibinfo{person}{Elliott Sober}.}
  \bibinfo{year}{2003}\natexlab{}.
\newblock \showarticletitle{Two uses of unification}.
\newblock In \bibinfo{booktitle}{\emph{The Vienna Circle and logical
  empiricism}}, \bibfield{editor}{\bibinfo{person}{Friedrich Stadler}} (Ed.).
  \bibinfo{publisher}{Kluwer Academic Publishers}, \bibinfo{address}{New York},
  \bibinfo{pages}{205--216}.
\newblock


\bibitem[Star(1989)]%
        {star1989structure}
\bibfield{author}{\bibinfo{person}{Susan~Leigh Star}.}
  \bibinfo{year}{1989}\natexlab{}.
\newblock \showarticletitle{The structure of ill-structured solutions: Boundary
  objects and heterogeneous distributed problem solving}.
\newblock In \bibinfo{booktitle}{\emph{Distributed artificial intelligence}}.
  \bibinfo{publisher}{Elsevier}, \bibinfo{pages}{37--54}.
\newblock


\bibitem[Thoma(2015)]%
        {thoma2015epistemic}
\bibfield{author}{\bibinfo{person}{Johanna Thoma}.}
  \bibinfo{year}{2015}\natexlab{}.
\newblock \showarticletitle{The epistemic division of labor revisited}.
\newblock \bibinfo{journal}{\emph{Philosophy of Science}} \bibinfo{volume}{82},
  \bibinfo{number}{3} (\bibinfo{year}{2015}), \bibinfo{pages}{454--472}.
\newblock


\bibitem[Varshney et~al\mbox{.}(2019)]%
        {varshney2019pretrained}
\bibfield{author}{\bibinfo{person}{Lav~R Varshney},
  \bibinfo{person}{Nitish~Shirish Keskar}, {and} \bibinfo{person}{Richard
  Socher}.} \bibinfo{year}{2019}\natexlab{}.
\newblock \showarticletitle{Pretrained AI models: performativity, mobility, and
  change}.
\newblock \bibinfo{journal}{\emph{arXiv preprint arXiv:1909.03290}}
  (\bibinfo{year}{2019}).
\newblock


\bibitem[Vashishth et~al\mbox{.}(2019)]%
        {vashishth2019attention}
\bibfield{author}{\bibinfo{person}{Shikhar Vashishth}, \bibinfo{person}{Shyam
  Upadhyay}, \bibinfo{person}{Gaurav~Singh Tomar}, {and}
  \bibinfo{person}{Manaal Faruqui}.} \bibinfo{year}{2019}\natexlab{}.
\newblock \showarticletitle{Attention interpretability across NLP tasks}.
\newblock \bibinfo{journal}{\emph{arXiv preprint arXiv:1909.11218}}
  (\bibinfo{year}{2019}).
\newblock


\bibitem[Vaswani et~al\mbox{.}(2017)]%
        {vaswani2017attention}
\bibfield{author}{\bibinfo{person}{Ashish Vaswani}, \bibinfo{person}{Noam
  Shazeer}, \bibinfo{person}{Niki Parmar}, \bibinfo{person}{Jakob Uszkoreit},
  \bibinfo{person}{Llion Jones}, \bibinfo{person}{Aidan~N Gomez},
  \bibinfo{person}{{\L}ukasz Kaiser}, {and} \bibinfo{person}{Illia
  Polosukhin}.} \bibinfo{year}{2017}\natexlab{}.
\newblock \showarticletitle{Attention is all you need}.
\newblock \bibinfo{journal}{\emph{Advances in Neural Information Processing
  Systems}}  \bibinfo{volume}{30}.
\newblock


\bibitem[Whittaker(2021)]%
        {whittaker2021steep}
\bibfield{author}{\bibinfo{person}{Meredith Whittaker}.}
  \bibinfo{year}{2021}\natexlab{}.
\newblock \showarticletitle{The steep cost of capture}.
\newblock \bibinfo{journal}{\emph{Interactions}} \bibinfo{volume}{28},
  \bibinfo{number}{6} (\bibinfo{year}{2021}), \bibinfo{pages}{50--55}.
\newblock


\bibitem[Wolpert and Macready(1997)]%
        {wolpert1997no}
\bibfield{author}{\bibinfo{person}{David~H Wolpert} {and}
  \bibinfo{person}{William~G Macready}.} \bibinfo{year}{1997}\natexlab{}.
\newblock \showarticletitle{No free lunch theorems for optimization}.
\newblock \bibinfo{journal}{\emph{IEEE transactions on evolutionary
  computation}} \bibinfo{volume}{1}, \bibinfo{number}{1}
  (\bibinfo{year}{1997}), \bibinfo{pages}{67--82}.
\newblock


\bibitem[Xiao et~al\mbox{.}(2021)]%
        {xiao2021training}
\bibfield{author}{\bibinfo{person}{Mingqing Xiao}, \bibinfo{person}{Qingyan
  Meng}, \bibinfo{person}{Zongpeng Zhang}, \bibinfo{person}{Yisen Wang}, {and}
  \bibinfo{person}{Zhouchen Lin}.} \bibinfo{year}{2021}\natexlab{}.
\newblock \showarticletitle{Training Feedback Spiking Neural Networks by
  Implicit Differentiation on the Equilibrium State}.
\newblock \bibinfo{journal}{\emph{Advances in Neural Information Processing
  Systems}}  \bibinfo{volume}{34} (\bibinfo{year}{2021}).
\newblock


\bibitem[Zollman(2010)]%
        {zollman2010epistemic}
\bibfield{author}{\bibinfo{person}{Kevin~JS Zollman}.}
  \bibinfo{year}{2010}\natexlab{}.
\newblock \showarticletitle{The epistemic benefit of transient diversity}.
\newblock \bibinfo{journal}{\emph{Erkenntnis}} \bibinfo{volume}{72},
  \bibinfo{number}{1} (\bibinfo{year}{2010}), \bibinfo{pages}{17--35}.
\newblock


\end{thebibliography}


\end{document}